%% file: main.tex
\documentclass[]{jingdong}
\microtypesetup{expansion=false}
\input{math_commands.tex}

\usepackage{url}
\usepackage{amsmath}
\usepackage{amssymb}
\usepackage{mathtools}
\usepackage{amsthm}
\usepackage{wrapfig}
\usepackage{array}
\usepackage{longtable}
\newlength{\originaltextfloatsep}
\usepackage{fvextra}

\DefineVerbatimEnvironment{PromptBox}{Verbatim}{
  breaklines=true,
  breakanywhere=true,
  breaksymbolleft={},
  breaksymbolright={},
  fontsize=\footnotesize,
  frame=single,
  framesep=3mm,
  rulecolor=\color{black!25},
  bgcolor=black!3
}

\title{EcomWriteBench: A Real-World Benchmark for E-commerce Writing Tasks}

\author[*]{Kaicheng Liang}
\author[*,\dagger]{Jingyi Cui}
\author{Hongwei Wen}
\author{Feng Wang}
\author{Simiu Gu}
\affiliation{JD.com}
\contribution[*]{Equal contribution}
\contribution[\dagger]{Corresponding author}
\correspondence{Jingyi Cui (\email{cuijingyi.12@jd.com}, \email{rubycui0113@gmail.com})}

\abstract{
Large language models (LLMs) are increasingly used throughout e-commerce workflows to generate content that informs purchasing decisions and supports customer interactions. Evaluating these systems requires accounting for user needs, product evidence, structural constraints, and business requirements that are not fully captured by general-purpose writing benchmarks. We introduce \textit{EcomWriteBench}, a real-world benchmark containing 3,861 queries from 69 e-commerce writing tasks across 13 categories and five languages. Through task-level extraction, cross-task consolidation, and refinement guided by human feedback, EcomWriteBench constructs a shared pool of 37 abilities and identifies the subset applicable to each task. Ability-wise scoring enables us to compare state-of-the-art LLMs on EcomWriteBench and characterize their relative strengths and weaknesses across fine-grained e-commerce writing abilities. We further propose a preference-calibrated weight-learning framework that converts human pairwise preferences into interpretable scoring rubrics through global, category-level, and task-level aggregation weights. This framework provides a reusable approach for translating human preferences into automated evaluation by combining ability discovery, ability-wise scoring, and preference-calibrated aggregation, and can potentially be applied to a broad range of benchmarks beyond EcomWriteBench.
{On held-out queries, our scoring framework achieves 88.96\% agreement with human preferences, compared with 78.54\% for conventional LLM-as-a-judge scoring.}
}

\begin{document}
\raggedbottom

\maketitle
\setlength{\textfloatsep}{10pt plus 2pt minus 2pt}

\section{Introduction}

Large language models (LLMs) have made rapid progress in open-domain generation and instruction following. E-commerce writing poses a distinct evaluation challenge because production tasks combine language generation with factual, structural, operational, and policy requirements~\cite{liu2026oxyecombench,wang2025ecom}. Marketing copy must communicate product information while satisfying style and compliance constraints; customer-service responses must be grounded in the available context and provide appropriate resolutions; logistics communication must accurately convey events and next steps. These requirements vary substantially across workflows and cannot be fully characterized by general measures of writing quality.

General-purpose writing benchmarks provide limited coverage of production e-commerce workflows, while a single holistic score or pairwise preference alone cannot show which task requirements a response satisfies. Recent work has expanded evaluation toward realistic writing and domain-specific benchmarks~\citep{wu2026writingbench,liu2026oxyecombench}, fine-grained rubric-based assessment~\citep{ye2024flask,kim2024prometheus,wei2025rocketeval}, and human-calibrated automatic metrics~\citep{winata2025metametrics,ryan2026autometrics}. These advances motivate an evaluation framework that jointly covers diverse e-commerce writing workflows, characterizes model behavior at the ability level, and converts human preferences into interpretable task-specific scoring functions.

We introduce \textit{EcomWriteBench}, a benchmark and evaluation framework for real-world e-commerce writing.
Its 3,861 multilingual queries span 69 tasks in 13 categories and five languages, including product-content generation, recommendation and decision support, customer service, logistics, and merchant operations.
The queries capture the context, output requirements, and constraints needed to judge a response within its task.

\begin{figure}[!t]
\centering
\vspace{-5mm}
\includegraphics[width=\linewidth]{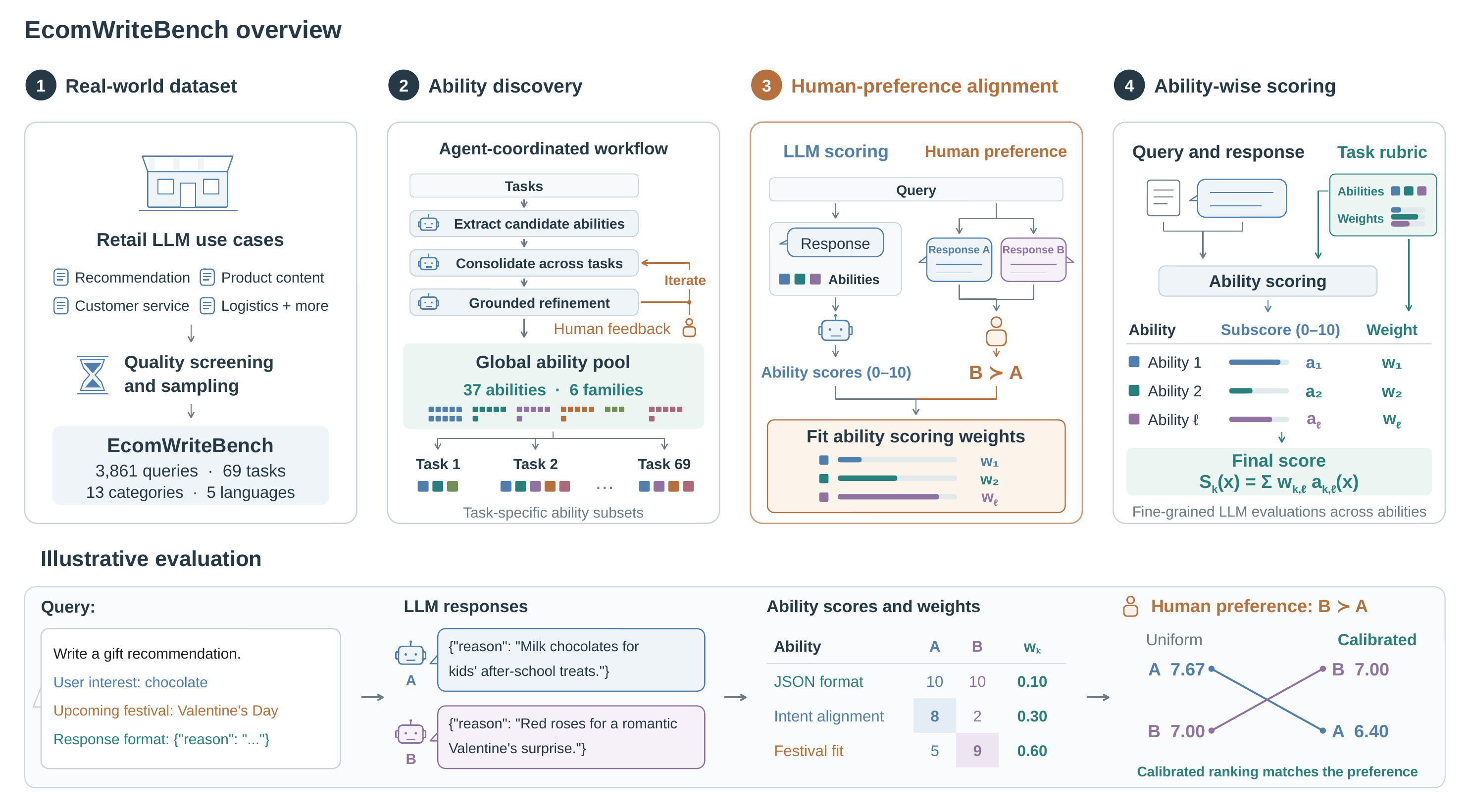}
\vspace{-5mm}
\caption{Overview of EcomWriteBench.
The framework combines benchmark construction from real-world retail tasks, iterative ability discovery, and scoring weights calibrated from human pairwise preferences.
The illustrative evaluation example shows how these weights aggregate fine-grained ability scores to match human preference despite competing strengths across abilities.
}
\vspace{-3mm}
\label{fig::pipeline}
\end{figure}

To support both diagnosis and overall model comparison, we construct task-adaptive rubrics and then apply them to model responses (Figure~\ref{fig::pipeline}).
Rubric construction begins with \textit{ability discovery}: we extract candidate abilities from individual tasks, consolidate them across tasks, and refine and supplement the shared pool through human feedback.
The resulting vocabulary contains 37 abilities, with a relevant subset for each task.
For calibration, we score the responses in human-annotated comparison pairs on these abilities and use the preferences to learn global weights with category- and task-level refinements.
At evaluation time, responses are scored on their task-relevant abilities, and the fixed learned weights combine those scores into an overall score.
The underlying ability scores remain available to explain each model's strengths and weaknesses.

Our experiments ask whether this scoring framework distinguishes model behavior and reflects human judgment.
Across 15 LLM configurations, we report preference-calibrated overall scores alongside unweighted profiles for six ability families; a 37-ability breakdown in Appendix~\ref{sec:fine_ability} identifies more specific strengths and weaknesses.
Models with similar overall scores can differ substantially in their ability profiles.
We also find that the learned weights vary across categories and tasks.
{On held-out queries, the full rubric agrees with human preferences more often than uniform ability aggregation, conventional LLM-as-a-judge scoring, and four recent human-calibrated evaluation methods.
The learned weights also generalize to unseen tasks, where they outperform uniform aggregation.}

\begin{itemize}

\item \textbf{Real-world e-commerce writing benchmark.}
We introduce \textit{EcomWriteBench}, containing 3,861 multilingual queries from 69 real-world e-commerce writing tasks across 13 categories and five languages. The benchmark covers diverse production workflows including recommendation, product-content generation, customer service, logistics, merchant operations, and post-purchase interaction. Together with the query collection, EcomWriteBench provides a fine-grained rubric-construction pipeline for task-adaptive evaluation, combining ability discovery, ability-wise scoring, and preference-calibrated aggregation.

\item \textbf{Ability discovery and ability-wise evaluation.}
We introduce an ability-discovery procedure that builds a shared vocabulary of 37 fine-grained abilities through task-level extraction, cross-task consolidation, and refinement guided by human feedback, and identifies the subset relevant to each task. Ability-wise scoring evaluates responses independently on these task-relevant dimensions, providing interpretable diagnostic signals and enabling consistent comparison of shared abilities across heterogeneous tasks. We provide the complete ability definitions, ability-discovery specifications, and scoring prompts to support reproducible and extensible ability-wise evaluation.\footnote{The benchmark data and evaluation code will be publicly released upon acceptance.}

\item \textbf{Preference-calibrated rubric learning via constrained optimization.}
We innovatively formulate rubric construction as an inverse preference-calibration problem and solve it with constrained optimization. Given human pairwise preferences over model responses, the optimization recovers interpretable global, category-level, and task-level weights over the discovered abilities. Combined with ability discovery and ability-wise scoring, the method provides a reusable pipeline for constructing fine-grained evaluation rubrics from preference annotations. Empirically, the learned rubrics achieve higher agreement with human preferences than uniform ability aggregation and heuristic LLM judging.

\end{itemize}

\FloatBarrier

\section{Related Work}
\label{sec:related-work}

\paragraph{Writing and Domain-Specific LLM Benchmarks.}
WritingBench~\citep{wu2026writingbench} evaluates writing across domains using query-dependent criteria.
FollowBench~\citep{jiang2024followbench} tests individual constraints as instructions grow more complex, while CFBench~\citep{zhang2025cfbench} assesses realistic constraint combinations and their prioritization.
WildBench~\citep{lin2025wildbench} evaluates challenging real-user queries with task-specific checklists and structured explanations.
Among e-commerce benchmarks, OxyEcomBench~\citep{liu2026oxyecombench} covers platform operators, merchants, and customers with text-only and multimodal inputs; ECom-Bench~\citep{wang2025ecom} studies customer-support agents; and EComStage~\citep{zhao2026ecomstage} evaluates agent-capable LLMs across perception, planning, and action stages.
EcomWriteBench focuses on writing-intensive production workflows with task-specific, ability-wise evaluation.

\paragraph{Fine-Grained LLM Evaluation and LLM-as-a-Judge.}
G-Eval~\citep{liu2023g} uses explicit evaluation criteria, while MT-Bench and Chatbot Arena~\citep{zheng2023judging} establish LLM-as-a-judge as a scalable evaluation paradigm.
FLASK~\citep{ye2024flask} evaluates instruction-dependent alignment skills, Prometheus~\citep{kim2024prometheus} supports customized rubrics, and LMUNIT~\citep{saad2025lmunit} scores natural-language unit tests with a trained evaluator.
EvalGen~\citep{shankar2024validates} supports human refinement of criteria and selects code- or prompt-based evaluators using human grades.
EcomWriteBench uses a shared vocabulary of task-grounded abilities and task-specific subsets for consistent ability-wise comparison across writing tasks.

\paragraph{Preference-Based Evaluation and Rubric Learning.}
Pairwise preferences provide supervision for reward learning~\citep{ouyang2022training}.
LLM-Rubric~\citep{hashemi2024llm} combines multidimensional judgments through a neural aggregator calibrated to individual annotators.
MetaMetrics~\citep{winata2025metametrics} combines existing metrics, RocketEval~\citep{wei2025rocketeval} decomposes evaluation into instance-specific checklist items and reweights them, and AutoMetrics~\citep{ryan2026autometrics} generates and combines task-specific metrics.
HD-Eval~\citep{liu2024hd} iteratively decomposes criteria and learns a preference-guided aggregator, while SAJA~\citep{kola2026saja} calibrates multidimensional rubric features with a lightweight head.
Our constrained hierarchical optimization learns interpretable global, category-level, and task-specific ability weights from pairwise preferences.

\section{EcomWriteBench}
\label{sec:ecomwritebench}

EcomWriteBench combines a curated collection of real-world e-commerce writing
queries with a task-adaptive evaluation rubric.
The rubric evaluates each response along a set of fine-grained abilities and
aggregates the resulting ability-wise scores using task-specific weights
calibrated from human pairwise preferences.
This construction provides both an overall score for model comparison and an
explicit decomposition of response quality across the abilities relevant to
each task.
Figure~\ref{fig::pipeline} provides an overview of the framework.

\subsection{Benchmark Construction}
\label{sec:benchmark-construction}

\paragraph{Task normalization and taxonomy.}
We construct EcomWriteBench from query pools spanning real-world e-commerce workflows with different prompt structures, output formats, and generation requirements.
We separate workflows with distinct objectives or output constraints into
individual tasks, yielding $K=69$ tasks organized into 13 functional
categories.

\begin{figure*}[!htb]
    \centering
    \begin{subfigure}[t]{0.63\linewidth}
        \centering
        \includegraphics[width=\linewidth]
        {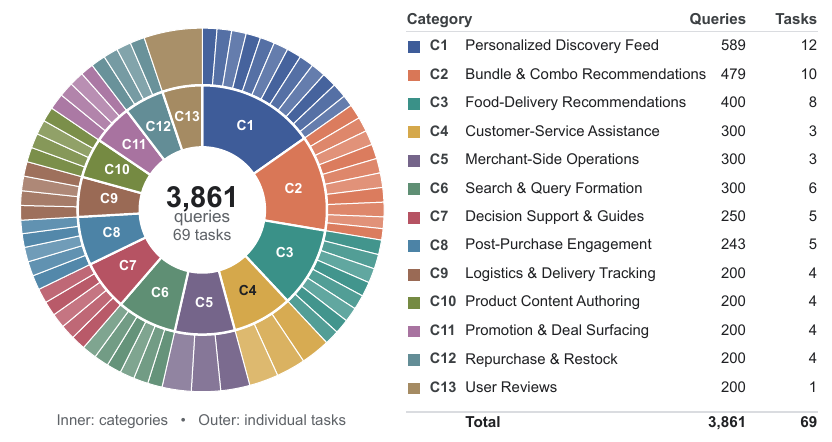}
        \caption{Task distribution}
        \label{fig:task-distribution}
    \end{subfigure}
    \hfill
    \begin{subfigure}[t]{0.30\linewidth}
        \centering
        \includegraphics[width=\linewidth]
        {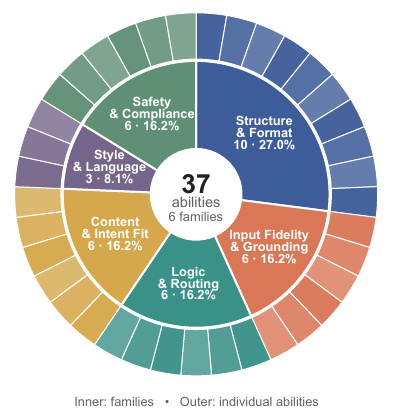}
        \caption{Ability pool}
        \label{fig:ability-pool}
    \end{subfigure}

    \caption{
    Overview of the EcomWriteBench task and ability composition.
    Left: distribution of the 3,861 benchmark queries across 13 task
    categories and 69 individual tasks, where the inner ring shows
    category-level composition and the outer ring shows the corresponding
    tasks.
    Right: composition and task coverage of the 37-ability evaluation pool
    obtained through task-level extraction, cross-task consolidation,
    and refinement guided by human feedback.
    The abilities are grouped into six families for
    visualization, and each ability is annotated with the number of benchmark
    tasks in which it is used.
    }
    \label{fig:ecomwritebench-composition}
\end{figure*}

\paragraph{Quality screening and sampling.}
Candidate queries undergo privacy-preserving rewriting to replace sensitive or organization-specific details while retaining their task requirements.
We screen out queries unsuitable for discriminative evaluation, using preliminary quality diagnostics and empty-response checks.
We then select queries using task-specific quotas to limit concentration in high-volume workflows while preserving coverage across categories.
The resulting benchmark contains 3,861 queries from all 69 tasks, with at least 200 queries per category.
Figure~\ref{fig:task-distribution} shows the composition of the final benchmark.
The queries are distributed approximately evenly across Chinese, English,
German, French, and Dutch, with task composition kept approximately aligned
across languages.
Additional details on multilingual composition, query lengths, and the
curation procedure are provided in Appendix~\ref{sec:dataset-details}.

\subsection{Ability Discovery}
\label{sec:ability-discovery}

We construct the shared ability vocabulary through a coordinated multi-agent workflow grounded in the requirements of individual e-commerce writing tasks.
Task-level agents extract candidate sub-abilities from task specifications and representative examples, identifying concrete requirements and criteria for judging whether they are satisfied.
The workflow then consolidates recurring requirements across tasks, merges redundant candidates, and examines their applicability.
The ability definitions and task assignments are iteratively refined through agent-based audits and human review, with domain-informed feedback guiding revisions to granularity, coverage, and the boundaries between related abilities.
Human feedback also supplements the pool with general writing abilities that are not captured by the initial task-level extraction.
The phased discovery procedure and prompt specifications are described in Appendix~\ref{sec:ability-discovery-details}.

The resulting vocabulary contains 37 abilities, denoted by $\mathcal{A}$.
Each task $T_k$ uses an applicable subset $\mathcal{A}_k=\{A_{k,1},\ldots,A_{k,M_k}\}\subseteq\mathcal{A}$, which specifies its $M_k$ evaluation dimensions.
The abilities are organized into six functional families for reporting and grouped judging in final evaluation.
Scores and learned weights remain ability-specific.
Figure~\ref{fig:ability-pool} shows the pool composition and task coverage; Appendix~\ref{sec:full-ability-inventory} provides the complete definitions, and Appendix~\ref{sec:judge-details} describes the scoring protocol.
We will provide a reusable agent skill based on this workflow to support rubric construction for other task collections.

\subsection{Ability-Wise Scoring}
\label{sec:llm-scoring}

For a model response $x$ to query $q$ from task $T_k$, we obtain a separate score for each applicable ability in $\mathcal{A}_k$.
For LLM-evaluated abilities, we aggregate scores from three heterogeneous judges to mitigate self-preference bias and improve evaluation stability and reliability~\citep{panickssery2024llm,verga2024replacing}.
Each judge assesses the response against the same ability definition and assigns an integer score $a_{k,\ell}^{(j)}(x)\in\{0,\ldots,10\}$ with a short justification.
We average the three judge scores to obtain
\begin{equation}
a_{k,\ell}(x)
=
\frac{1}{3}
\sum_{j=1}^{3}
a_{k,\ell}^{(j)}(x),
\label{eq:ability-scoring}
\end{equation}
and use the programmatic score on the same 0--10 scale for $a_{k,\ell}(x)$ on deterministically checked abilities.
We collect the applicable ability scores into
\begin{equation}
\boldsymbol{a}_k(x)
=
[a_{k,1}(x),\ldots,a_{k,M_k}(x)]^\top.
\end{equation}

\subsection{Preference-Calibrated Rubric Learning}
\label{sec:weight-learning}

The relative importance of rubric dimensions varies across e-commerce writing
tasks.
We learn their aggregation weights from human pairwise preferences.
{The annotators are retail-industry practitioners, and only pairs whose annotations show a clear preference pattern are retained.}
For task $T_k$, let
$\mathcal{P}_k=\{(x_i,x_j):x_i\prec x_j\}$
denote the available preference pairs, where $x_j$ is preferred to $x_i$.
For pair $p=(x_i,x_j)$, let $\bar{\boldsymbol{\Delta}}_p$ denote the raw score difference $\boldsymbol{a}_k(x_j)-\boldsymbol{a}_k(x_i)$ embedded in the shared ability space, with zeros for inapplicable abilities.
The optimization uses the variance-scaled difference $\Delta_{p,\ell}=\bar{\Delta}_{p,\ell}/d_\ell$, where the scale factors $d_\ell$ are defined in Appendix~\ref{sec:optimization-details}.

The learned weights are optimized so that the weighted ability-wise rubric
assigns higher scores to human-preferred responses.

\paragraph{Hierarchical weight sharing.}
To share varying amounts of task supervision, we use the hierarchy
$\boldsymbol{w}^{G}
\rightarrow
\boldsymbol{w}^{C}_{c}
\rightarrow
\boldsymbol{w}^{T}_{k}$,
where $\boldsymbol{w}^{G}$ captures ability importance shared across all
tasks, $\boldsymbol{w}^{C}_{c}$ adapts the aggregation rule to category $c$,
and $\boldsymbol{w}^{T}_{k}$ further specializes the rubric to task $T_k$.

Let $\mathcal{T}_c$ denote the set of tasks in category $c$, and let $c(k)$
denote the category containing task $T_k$.
When a global or category-level vector is applied to task $T_k$, only
components corresponding to abilities in $\mathcal{A}_k$ are used.

\paragraph{L2-regularized formulation.}
The global fit pools training pairs across tasks; category and task fits use their corresponding subsets.
The levels are fitted sequentially, keeping each fitted parent fixed when optimizing its children.
Each retained preference pair contributes equally to the summed loss at the level where it is used.
The objectives are
\begin{equation}
\begin{aligned}
\boldsymbol{w}^{G}
&=
\arg\min_{\boldsymbol{w}}
\sum_{k}
\sum_{p\in\mathcal{P}_k}
\left[
m-\boldsymbol{w}^{\top}\boldsymbol{\Delta}_p
\right]_{+}^{2},
\\
\boldsymbol{w}^{C}_{c}
&=
\arg\min_{\boldsymbol{w}}
\sum_{k\in\mathcal{T}_c}
\sum_{p\in\mathcal{P}_k}
\left[
m-\boldsymbol{w}^{\top}\boldsymbol{\Delta}_p
\right]_{+}^{2}
+
\lambda
\left\|
\boldsymbol{w}-\boldsymbol{w}^{G}
\right\|_{2}^{2},
\\
\boldsymbol{w}^{T}_{k}
&=
\arg\min_{\boldsymbol{w}}
\sum_{p\in\mathcal{P}_k}
\left[
m-\boldsymbol{w}^{\top}\boldsymbol{\Delta}_p
\right]_{+}^{2}
+
\mu
\left\|
\boldsymbol{w}-\boldsymbol{w}^{C}_{c(k)}
\right\|_{2}^{2},
\end{aligned}
\label{eq:l2-weight-learning}
\end{equation}
where $[z]_+=\max(z,0)$ and $m=1$ is the ranking margin.
All levels optimize the full shared ability vector subject to $w_\ell\geq0$.
These constraints apply in the variance-scaled coefficient space; inapplicable ability features are zero.
The coefficient $\lambda$ controls how strongly category-level weights are pulled toward the global weights, while $\mu$ controls how strongly task-specific weights are pulled toward their category-level weights.
Larger values favor greater weight sharing; a value of zero removes the corresponding sharing penalty.
There is no independent norm penalty or proportional cap; normalization is applied only when converting the fitted coefficients into task-specific scoring weights.

We use the L2 formulation as the primary setting.
An L1 counterpart using slack variables and L1 hierarchical penalties is
provided in Appendix~\ref{sec:l1-formulation}.

\subsection{Final Scoring Rule}
\label{sec:final-scoring}

The final EcomWriteBench rubric combines the ability-wise scoring representation
with the preference-calibrated task-specific aggregation weights.
For a response $x$ from task $T_k$, only the abilities in
$\mathcal{A}_k$ are evaluated using the protocol in
Section~\ref{sec:llm-scoring}.
The ability-scoring protocol is detailed in Appendix~\ref{sec:judge-details}.

Let $\boldsymbol{w}_k$ denote the final task-specific aggregation vector
obtained by back-transforming $\boldsymbol{w}_k^T$ to the raw score scale and normalizing over the abilities applicable to task $T_k$, such that
$\sum_{\ell=1}^{M_k}w_{k,\ell}=1$ and $w_{k,\ell}\geq0$.
The EcomWriteBench rubric score for response $x$ is
\begin{equation}
S_k(x)
=
\boldsymbol{w}_k^\top\boldsymbol{a}_k(x)
=
\sum_{\ell=1}^{M_k}
w_{k,\ell}a_{k,\ell}(x).
\label{eq:final-task-score}
\end{equation}

The score is therefore a task-specific weighted aggregation of independently
evaluated ability dimensions.
The ability-wise scores characterize how well the response satisfies each
rubric dimension, while the learned weights encode the relative importance of
these dimensions according to human pairwise preferences.
Abilities outside $\mathcal{A}_k$ do not contribute to $S_k(x)$.

For benchmark-level model comparison, let
$\mathcal{X}_{g,k}$ denote the responses produced by model $g$ for task
$T_k$.
We average rubric scores over all evaluation queries, giving each query equal weight:
\begin{equation}
S_g
=
\frac{1}{\sum_{k=1}^{K}|\mathcal{X}_{g,k}|}
\sum_{k=1}^{K}
\sum_{x\in\mathcal{X}_{g,k}}
S_k(x).
\label{eq:overall-benchmark-score}
\end{equation}
Consequently, each task contributes in proportion to its number of evaluation queries.

\section{Experiments}
\label{sec:experiments}

We first examine how the learned rubric weights vary across tasks, then evaluate their agreement with pairwise preferences, and finally use the calibrated rubric to compare models' e-commerce writing capabilities.

\subsection{Experimental Settings}
\label{sec:experiment-settings}

We evaluate the 15 LLM configurations listed in Table~\ref{tab:model-family-main} on the full EcomWriteBench benchmark.
We use Gemini-3.1-Pro, Gemma-4-31B, and Qwen3.6-35B-A3B as LLM judges.
For LLM-evaluated abilities, we use the mean of their scores; abilities covered by deterministic checks are scored programmatically.
Judge configurations and prompts are given in Appendix~\ref{sec:judge-details}, with judge-consistency analyses in Appendix~\ref{sec:ability-judge-reliability}.

The main experiments use the L2 formulation and the \emph{strict preference set}, retaining only pairs whose annotations unanimously favor one response.
Alternative preference-selection criteria and the L1 formulation are evaluated in Appendix~\ref{sec:l1-formulation}.

\subsection{Learned Rubric Weights}
\label{sec:learned-rubric-weights}

\begin{figure}[htb]
    \centering
    \includegraphics[
        width=0.88\linewidth
    ]{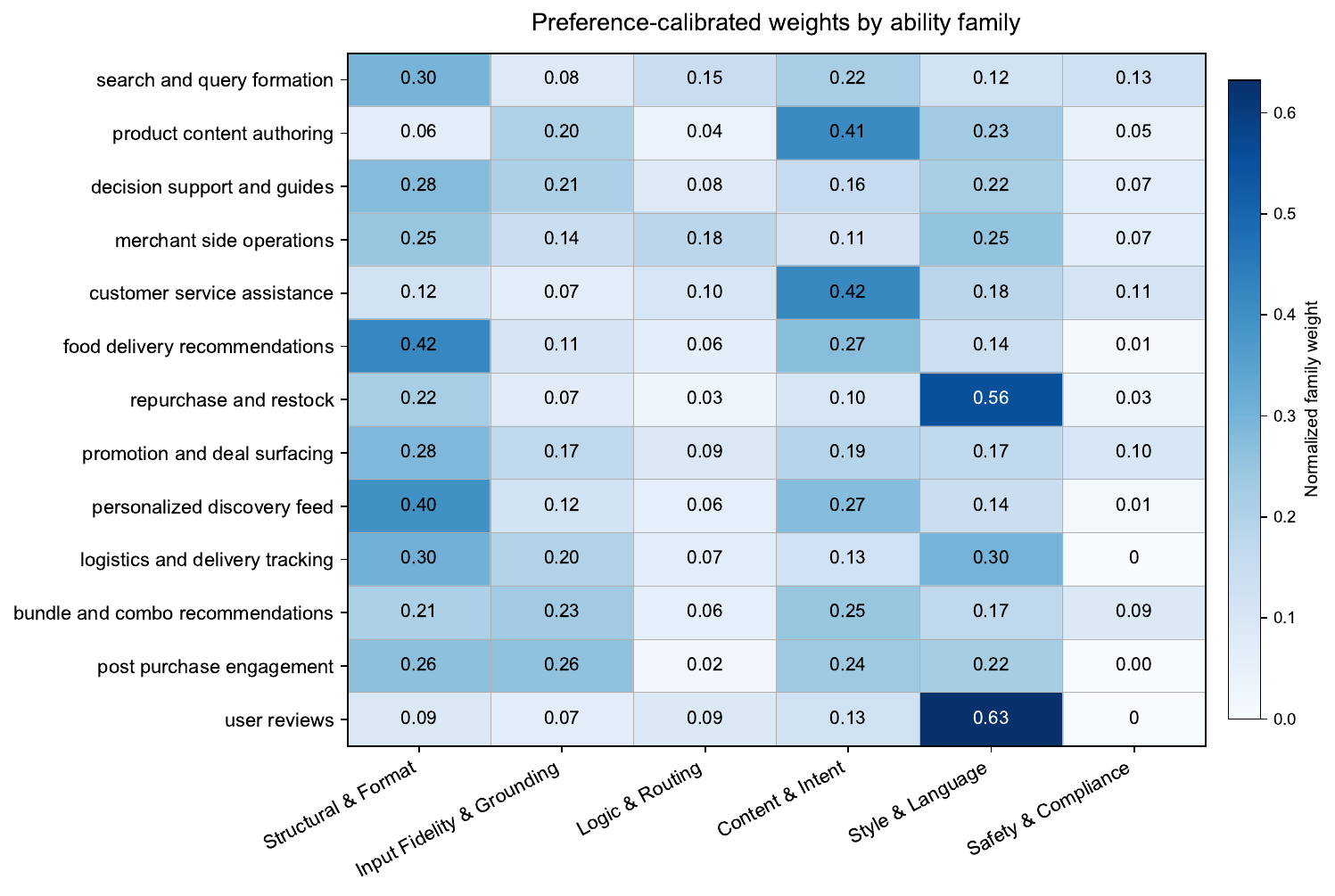}
    \caption{
    Category-level preference-calibrated rubric weights aggregated over the
    six ability families.
    Each row is normalized to sum to one.
    Darker cells indicate larger relative weights.
    }
    \label{fig:category-family-weights}
\end{figure}

We examine how the learned rubric weights vary across e-commerce tasks.
Following the hierarchy in Section~\ref{sec:weight-learning}, global weights are fitted to pooled preferences across tasks, category-level weights to preferences from tasks in the same category, and task-specific weights to preferences from the target task.
Figure~\ref{fig:category-family-weights} shows how each category's total weight is distributed across the six ability families.
Each family weight is the sum of its ability weights, and the six family weights in a category are normalized to sum to one.

The profiles differ across categories.
\emph{Content \& Intent} receives the largest share in product content authoring and customer service assistance, while \emph{Style \& Language} dominates user reviews and repurchase and restock.
\emph{Structural \& Format} receives the largest share in food delivery recommendations and personalized discovery feed.
These profiles show different learned priorities across task contexts.
{The differences follow the nature of the tasks.
Product content and customer service replies are judged first on whether they address the user's actual need, so \emph{Content \& Intent} dominates.
In user reviews and repurchase reminders, the input largely determines what to say, so responses differ mainly in how naturally they read.
Food delivery and personalized discovery feeds are rendered in fixed display formats, so \emph{Structural \& Format} dominates.}

The full task-specific weight matrix in Appendix~\ref{sec:full-task-rubric-weights} shows further variation among tasks within the same category.
We next evaluate the preference alignment of the resulting rubric and the benefit of task-specific calibration.

\subsection{Preference Alignment of Learned Weights}
\label{sec:weight-results}

{We first compare the complete rubric with alternative scoring schemes and four human-calibrated evaluation methods, then assess weight granularity on observed tasks and transfer to unseen tasks.}
Pairwise agreement is the fraction of evaluated pairs for which the preferred response receives a higher score; score ties count as incorrect.

\paragraph{Validation protocol.}
{All preference-fitted scoring functions are evaluated on held-out queries or unseen tasks.}
For \emph{observed tasks}, we split the queries within each task into five folds, keeping all response comparisons for the same query in the same fold.
In each round, four folds are used to fit the weights and the remaining fold is used for evaluation.
Global, category-level, and task-specific weights use the same partitions.
Global weights have no hierarchical sharing parameter; category-level weights select $\lambda$, and task-specific weights select $(\lambda,\mu)$ by agreement pooled across the five validation folds.
The results report validation agreement at these selected settings.

For \emph{unseen tasks}, we leave out all preference pairs from one task at a time.
Global weights are fitted using all remaining tasks, while category-level weights are fitted using the other tasks in the held-out task's category.
We then apply each weight vector to the abilities relevant to the held-out task.
For the unseen-task comparison, we exclude the sole task in the user reviews category because no other task in that category remains available to fit category-level weights.
{On the remaining 68 tasks, we compare global and category-level transfer with uniform ability aggregation.}
The category-level $\lambda$ is selected by agreement pooled across these held-out tasks.

\paragraph{{Comparison with alternative scoring methods.}}
{Table~\ref{tab:scoring-scheme-ablation} compares the EcomWriteBench rubric with two parameter-free scoring schemes and four human-calibrated evaluation methods under the observed-task protocol above.}
\emph{Conventional LLM-as-a-judge scoring} serves as a baseline for prompt-based overall evaluation.
Each of the same three LLM judges assigns one holistic score using a task-specific prompt covering instruction following, correctness, faithfulness to the input, format constraints, and language quality.
Their scores are averaged, and the relative contribution of each criterion is left to the judge.
\emph{Uniform ability aggregation} averages the applicable ability scores with equal weights.
The \emph{EcomWriteBench rubric} combines those same ability scores with learned task-specific weights.
{We also reproduce LLM-Rubric~\citep{hashemi2024llm}, HD-Eval~\citep{liu2024hd}, MetaMetrics~\citep{winata2025metametrics}, and AutoMetrics~\citep{ryan2026autometrics} end to end, so that each method builds its own features and fits its own aggregator on the training folds.
These methods are designed to fit scalar quality ratings, whereas our supervision consists of pairwise preferences.
We therefore train each aggregator either on the feature difference between the two responses of a pair or on scalar Bradley--Terry strengths estimated from the training pairs, and report the better variant for each method.
We use the official implementations where available and follow the model choices of the original papers where possible. Details are given in Appendix~\ref{sec:baseline-details}.}

\begin{table}[htb]
\centering
\small

\caption{{
Held-out agreement with human preferences on observed tasks, using five-fold cross-validation with queries held out within each task.
Gain denotes the improvement of EcomWriteBench over each baseline in percentage points.
}}
\label{tab:scoring-scheme-ablation}
\begin{tabular}{lcc}
\toprule
Method & Agreement $\uparrow$ & Gain $\uparrow$ \\
\midrule
Conventional LLM-as-a-judge & 0.7854 & +10.42 pt \\
Uniform ability aggregation & 0.8252 & +6.44 pt \\
\midrule
AutoMetrics & 0.7406 & +14.90 pt \\
MetaMetrics & 0.7558 & +13.38 pt \\
HD-Eval & 0.8241 & +6.55 pt \\
LLM-Rubric & 0.8634 & +2.62 pt \\
\midrule
EcomWriteBench rubric & \textbf{0.8896} & -- \\
\bottomrule
\end{tabular}
\end{table}

{The EcomWriteBench rubric achieves the highest held-out agreement.
Uniform ability aggregation improves agreement by 3.98 percentage points over conventional LLM-as-a-judge scoring, which supports the task-grounded ability representation: the discovered abilities explicitly encode output requirements and support evaluation across the dimensions relevant to each task.
Learning task-specific weights adds another 6.44 percentage points, supporting task-specific weighting as a mechanism for aligning rubric scores with human judgments of response quality.
Among the reproduced methods, LLM-Rubric is the strongest at 0.8634, 2.62 percentage points below the EcomWriteBench rubric.
The final benchmark rubrics used for model evaluation are fitted on all preference pairs rather than on training folds and reach an in-sample agreement of 0.9042 with the annotators.}

\begin{table*}[htb]
\centering

\begin{minipage}[t]{0.48\textwidth}
\centering
\captionof{table}{
Pairwise agreement under different weight granularities on observed tasks.
Results use five-fold cross-validation with queries held out within each task.
}
\label{tab:weight_granularity2}

\vspace{1mm}

\begin{tabular}{lcc}
\hline
Weight granularity & Agreement & Gain \\
\hline
Global
& 0.8515
& -- \\

Category
& 0.8709
& +1.94 pt \\

Task
& \textbf{0.8896}
& +3.81 pt \\
\hline
\end{tabular}
\end{minipage}
\hfill
\begin{minipage}[t]{0.48\textwidth}
\centering
\captionof{table}{
{Generalization of shared aggregation weights to unseen tasks.
All rows use the same 68 held-out tasks and preference pairs; gains are relative to uniform ability aggregation.}
}
\label{tab:weight_granularity}

\vspace{1mm}

\begin{tabular}{lcc}
\hline
{Weights} & Agreement & Gain \\
\hline
{Uniform}
& {0.8259}
& {--} \\

Global
& 0.8464
& {+2.05 pt} \\

Category
& \textbf{0.8484}
& {+2.24 pt} \\
\hline
\end{tabular}

\end{minipage}

\end{table*}

\paragraph{Weight granularity on observed tasks.}
Preference agreement increases from 0.8515 with global weights to 0.8709 with category-level weights and 0.8896 with task-specific weights (Table~\ref{tab:weight_granularity2}).
The 3.81-percentage-point gain over global weighting supports learning a separate aggregation rule for each task when preference supervision is available.
{The same ordering holds when the ability scores come from any single judge (Appendix~\ref{sec:ability-judge-reliability}).}

\paragraph{Generalization to unseen tasks.}

{On the 68 held-out tasks, neither global nor category-level weights use any preference from the target task, yet both agree with human preferences more often than uniform ability aggregation (Table~\ref{tab:weight_granularity}).
Global weights reach 0.8464, 2.05 percentage points above the 0.8259 of uniform aggregation (95\% confidence interval $[1.16, 2.95]$ from a bootstrap over tasks), so the learned ability weighting transfers to tasks without preference annotations.
Category-level weights reach 0.8484; their 0.19-point margin over global weights is not statistically significant, which indicates that global weights already suffice for a new task and that finer granularity pays off mainly when preferences for the task are available.}

Implementation details and sensitivity analyses are provided in Appendices~\ref{sec:optimization-details}, \ref{sec:shrinkage-ablation}, and~\ref{sec:margin-ablation}.

\subsection{Overall and Ability-Family Comparison}
\label{sec:model-comparison}

We now apply the calibrated rubrics to compare model performance on EcomWriteBench.
Table~\ref{tab:model-family-main} reports overall scores computed using the preference-calibrated rubric in Section~\ref{sec:final-scoring}, together with scores for the six ability families.
Each family score averages the raw scores from all applicable response--ability evaluations within that family and summarizes performance in that ability category.

\begin{table*}[!htb]
\centering
\small
\caption{
Overall EcomWriteBench performance and ability-family scores of the 15 model
configurations included in the comparison.
Both overall and family scores are reported on the 0--10 scale.
Overall scores use the preference-calibrated rubric, whereas family scores
are unweighted means of the corresponding raw ability evaluations.
Bold indicates the best result among models not used as judges; $^\dagger$ marks the three judge configurations.
{The 95\% confidence intervals of the overall scores are obtained by resampling the 69 tasks with replacement 20,000 times.}
}
\label{tab:model-family-main}

\setlength{\tabcolsep}{3.2pt}
\begin{tabular}{llrcrrrrrr}
\toprule
& & \multicolumn{2}{c}{{Overall (0--10)}}
& \multicolumn{6}{c}{Ability-family score (0--10)} \\
\cmidrule(lr){3-4}\cmidrule(lr){5-10}
Model
& Reasoning
& {Score}
& {95\% CI}
& Struct.
& Fidelity
& Logic
& Content
& Style
& Safety \\
\midrule

GPT-5.6-Sol
& max
& \textbf{9.581}
& {[9.499, 9.660]}
& 9.64
& \textbf{9.70}
& \textbf{9.33}
& 9.30
& 9.33
& \textbf{9.84} \\

Gemini-3.1-Pro$^\dagger$
& high
& 9.577
& {[9.500, 9.653]}
& 9.67
& 9.55
& 9.14
& 9.33
& 9.53
& 9.61 \\

GLM-5.3
& max
& 9.539
& {[9.454, 9.624]}
& 9.24
& 9.47
& 9.25
& \textbf{9.56}
& \textbf{9.55}
& 9.70 \\

Kimi-K3
& max
& 9.535
& {[9.463, 9.608]}
& 9.21
& 9.50
& 9.22
& 9.55
& 9.47
& 9.72 \\

Claude-Opus-4.8
& max
& 9.414
& {[9.327, 9.503]}
& 9.05
& 9.31
& 9.04
& 9.38
& 9.48
& 9.72 \\

Gemma-4-31B$^\dagger$
& enabled
& 9.379
& {[9.285, 9.475]}
& 9.22
& 9.50
& 8.96
& 9.10
& 9.34
& 9.68 \\

Qwen3.8-27B
& xhigh
& 9.338
& {[9.230, 9.446]}
& \textbf{9.73}
& 9.62
& 9.05
& 8.86
& 8.69
& 9.83 \\

Claude-Sonnet-5
& max
& 9.313
& {[9.224, 9.406]}
& 9.12
& 9.28
& 8.97
& 9.17
& 9.24
& 9.63 \\

MiniMax-M3
& enabled
& 9.215
& {[9.056, 9.357]}
& 8.80
& 9.20
& 8.84
& 9.05
& 9.07
& 9.67 \\

Gemma-4-31B
& disabled
& 9.184
& {[9.066, 9.302]}
& 8.85
& 9.37
& 8.73
& 9.07
& 9.29
& 9.60 \\

Qwen3.5-397B-A17B
& enabled
& 9.180
& {[9.043, 9.314]}
& 9.44
& 9.28
& 8.88
& 8.82
& 8.71
& 9.66 \\

Qwen3.6-35B-A3B$^\dagger$
& enabled
& 9.179
& {[9.054, 9.307]}
& 9.30
& 9.11
& 8.87
& 8.95
& 8.83
& 9.59 \\

DeepSeek-V4-Pro
& max
& 8.961
& {[8.822, 9.100]}
& 8.53
& 8.72
& 8.14
& 8.97
& 9.32
& 9.44 \\

Qwen3.5-397B-A17B
& disabled
& 8.937
& {[8.780, 9.090]}
& 8.67
& 8.88
& 8.34
& 8.84
& 9.19
& 9.45 \\

DeepSeek-V4-Flash
& max
& 7.785
& {[7.438, 8.110]}
& 7.46
& 7.42
& 7.55
& 7.92
& 7.40
& 9.31 \\

\bottomrule
\end{tabular}
\end{table*}

\paragraph{Overall and family-level performance.}
{GPT-5.6-Sol, Gemini-3.1-Pro, GLM-5.3, and Kimi-K3 form the leading group, with overall scores between 9.535 and 9.581.
The gap between this group and the next configuration, Claude-Opus-4.8, is significant (0.122, 95\% confidence interval $[0.081, 0.164]$).}
Although these four models differ by less than 0.05 overall, their family profiles reveal distinct strengths: GPT-5.6-Sol leads on Fidelity, Logic, and Safety, whereas GLM-5.3 leads on Content and Style.

Strengths in individual families also differ from the overall ranking.
Qwen3.8-27B with reasoning achieves the highest Structural \& Format score (9.73), but scores 8.86 on Content and 8.69 on Style.
DeepSeek-V4-Pro scores 9.32 on Style, compared with 8.53 on Structural \& Format, 8.72 on Fidelity, and 8.14 on Logic.
These contrasts illustrate how the ability profiles help explain differences that an overall score alone cannot characterize.

\emph{Input-Conditioned Logic \& Routing} has the lowest mean across the 15 configurations (8.82) and is the lowest-scoring family for seven of them.
The weakest family nevertheless varies by model: GPT-5.6-Sol scores lowest on Content, GLM-5.3 and Kimi-K3 on Structural \& Format, and Qwen3.8-27B with reasoning on Style.
The full 37-ability breakdown in Appendix~\ref{sec:fine_ability} provides more detailed diagnostics within these families.
{Per-language results are reported in Appendix~\ref{sec:per-language}.}

\paragraph{Effect of reasoning.}
\label{sec:reasoning-analysis}
Enabling reasoning mainly improves abilities involved in satisfying task conditions and output constraints.
Comparisons between enabled and disabled configurations of Qwen3.5-397B-A17B and Gemma-4-31B show gains in input-variable use, interaction-history handling, instruction prioritization, JSON validity, and length control.
Improvements in content and style are less consistent, with minor fluctuations across some dimensions.

\section{Conclusion}

We introduce \textit{EcomWriteBench}, a multilingual benchmark for real-world e-commerce writing comprising 3,861 queries from 69 tasks across 13 categories and five languages, together with a methodology for constructing fine-grained, preference-calibrated evaluation rubrics. The benchmark covers diverse production workflows, while task-grounded ability discovery produces a shared vocabulary of 37 abilities and task-specific subsets for interpretable, ability-wise evaluation. We formulate aggregation-weight learning as constrained optimization over human pairwise preferences, yielding global, category-level, and task-specific weights that capture differences in evaluation priorities across tasks. Experiments across 15 LLM configurations reveal distinct ability profiles and differences in performance across e-commerce writing tasks. 
{On held-out queries, the learned rubrics agree with human preferences more often than uniform ability aggregation, conventional LLM-as-a-judge scoring, and four human-calibrated evaluation methods.}
The resulting framework combines shared evaluation dimensions with task-adaptive aggregation, enabling fine-grained analysis across heterogeneous tasks. For other task collections with suitable preference annotations, the methodology offers a reusable approach to constructing interpretable evaluation rubrics calibrated to human preferences.

\setlength{\textfloatsep}{\originaltextfloatsep}
\subsection*{AI use statement}

In this work, we used generative AI tools for generating model responses to benchmark queries collected from real-world e-commerce data, supporting qualitative analysis of task requirements for ability discovery and rubric development, performing automated scoring as part of the proposed evaluation method, implementing methods and experimental procedures, assisting with translation, and cleaning, reformatting, and privacy-preserving rewriting of data during dataset preparation. 

We have not used generative AI tools for conceptual framing, designing or providing feedback on research methodology or experiments, or interpreting experimental results, and tasks related to formulating mathematical claims, developing or assisting with mathematical proofs, and proposing research hypotheses are not applicable to this work. 

Additionally, we used generative AI tools, including Codex and Claude Code, for creating and editing software code, debugging and executing experimental scripts, computing and organizing experimental results, searching for related literature, assisting with manuscript drafting, grammar checking, and language polishing. 

We have reviewed all AI-assisted work. AI-generated or AI-assisted benchmark artifacts and processed data were reviewed by the authors before use; AI-assisted code and experimental scripts were checked and tested; and experimental results, references, and manuscript text were reviewed and verified by the authors. We take responsibility for the final content of this work, including text, claims or artifacts produced with the aid of generative AI.

\subsection*{Ethics statement}

Data use in this study was limited to academic research.
We applied de-identification measures to protect personal privacy and commercial confidentiality.
Pairwise preference annotation was carried out by e-commerce practitioners, who were informed of the research purpose and annotation procedures.

\subsection*{Reproducibility statement}

We provide detailed specifications of the EcomWriteBench evaluation protocol and preference-calibrated weight-learning procedure in Appendices~\ref{sec:judge-details} and~\ref{sec:optimization-details}, including the ability definitions, scoring prompts, judge configuration, and optimization details. The EcomWriteBench dataset, evaluation code, and implementation for learning ability weights from human preferences will be publicly released upon acceptance. We will additionally release reusable workflow aids for benchmark evaluation and rubric construction, covering input preparation, procedure configuration and execution, and adaptation to new datasets and evaluation requirements.

\bibliography{EcomWriteBench}
\bibliographystyle{iclr2027_conference}

\clearpage
\appendix
\section{Appendix}
\label{sec:appendix}

\subsection{Benchmark Construction and Dataset Details}
\label{sec:dataset-details}

\subsubsection{Task Normalization, Curation, and Sampling}

Each source workflow is represented using a common task specification that
records its input structure, expected output format, generation requirements,
and available query pool.
Workflows with different generation objectives or materially different output
constraints are treated as separate tasks.
This normalization produces the 69 tasks used throughout EcomWriteBench.

During privacy-preserving rewriting, personally identifying information,
organization-specific names, platform-specific terminology, and other
production-specific entities are replaced with generic or synthetic
alternatives.
The rewriting procedure preserves the information required to complete the
task, including task variables, factual dependencies, numerical constraints,
structured-output requirements, and contextual relationships.

{Validation} is applied after rewriting to identify malformed, degenerate, or privacy-sensitive instances.
Queries that do not retain sufficient information for reliable model
evaluation are excluded.
The resulting candidate corpus contains 54,403 queries across the 69 tasks.

Before sampling, we exclude queries flagged by preliminary quality diagnostics, including those with an empty response from any preliminary model.
Task-specific quotas limit concentration in high-volume workflows while retaining all 69 tasks.
Within each task, eligible queries are 
{sampled at random.}
Retaining all tasks also preserves coverage of their 37 assigned abilities.
The final benchmark contains 3,861 queries across 13 categories, with at least 200 queries in each category.

\subsubsection{Multilingual Composition}

The benchmark is distributed approximately evenly across five languages.
The exact numbers are 774 Chinese, 773 English, 772 German, 772 French,
and 770 Dutch queries, for a total of 3,861 queries.

Language assignment is performed within individual tasks so that task
composition remains approximately aligned across language subsets.
Translation preserves task variables, factual dependencies, and structured-output requirements.
{Every translation was verified by human reviewers for semantic fidelity and task transferability.}
Language-dependent length, character-class, and output-language constraints are adapted to the assigned language.
The ability definitions provided to evaluators remain in English across all
five language subsets.

\subsubsection{Query Length Statistics}

\begin{figure}[!h]
    \centering
    \includegraphics[width=\linewidth]{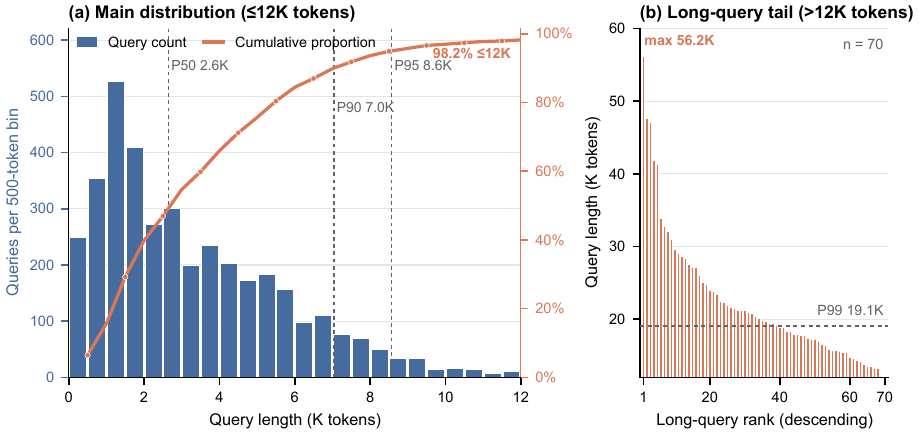}
    \caption{
    Token-length distribution of the 3,861 multilingual EcomWriteBench queries.
    The left panel shows the distribution up to 12K tokens and its cumulative proportion; the right panel shows the 70 longer queries.
    Approximate P50/P90/P95/P99 lengths are 2.6K/7.0K/8.6K/19.1K tokens.
    }
    \label{fig:token-distribution}
\end{figure}

The 3,861 multilingual benchmark queries contain 13,859,966 tokens in total, with a mean length of 3,590 tokens, measured on the query text using the Qwen3-30B-A3B tokenizer.
The median length is approximately 2.6K tokens, while the 90th, 95th, and 99th percentiles are approximately 7.0K, 8.6K, and 19.1K tokens, respectively.
As shown in Figure~\ref{fig:token-distribution}, 98.2\% of queries contain at most 12K tokens; 70 exceed this length, and the maximum is 56,230 tokens.

\subsubsection{Ability Discovery and Consolidation}
\label{sec:ability-discovery-details}

The discovery harness coordinates successive phases of candidate extraction, cross-task consolidation, and grounded refinement.
Within these phases, task-level extraction, assignment, and audit work is distributed across parallel agents.

\paragraph{Task-level candidate extraction.}
Agents read each task's shared instructions and representative examples to identify the requirements that its responses should satisfy.
Extraction prioritizes broad coverage rather than early compression, retaining potentially overlapping candidates for subsequent consolidation.
Each candidate includes a definition, the supporting task requirement, and a description of what a judge should inspect in the query and response to assess compliance.
This establishes a task-grounded candidate set before introducing cross-task abstractions.

\paragraph{Cross-task consolidation.}
Complementary consolidation proposals are synthesized into a shared ability pool and checked for coverage, clarity, and overlap.
The aim is to capture recurring requirements with reusable definitions while preserving distinctions needed for diagnosis.
Proposed merges are scrutinized for whether they obscure distinct failure modes or combine incompatible evaluation procedures; frequent co-occurrence alone is not sufficient evidence for merging.
Definitions are refined to specify both their scope and their boundaries with neighboring abilities, and the consolidated pool is mapped back to the tasks that exercise each ability.

\paragraph{Grounded refinement and feedback.}
The pool is then assessed through its application to real tasks: parallel assignment and audit passes examine task coverage and whether the definitions support scoring representative query--response pairs.
Findings are aggregated to distinguish incorrect assignments from missing or ambiguous criteria in the shared pool.
Human review and domain-informed feedback guide targeted revisions to the pool's granularity, definitions, and task applicability.
After substantive definition changes, the relevant tasks are reassessed against their underlying requirements rather than inheriting an earlier identifier mapping.
Repeated refinement and reassignment produce the final 37-ability vocabulary and its task-specific subsets.
The six functional families in Appendix~\ref{sec:full-ability-inventory} organize the finalized vocabulary and support the grouped evaluation described in Appendix~\ref{sec:judge-details}.

To facilitate reuse, we will release the ability-discovery procedure as an open-source agent skill for applying the pipeline to new task collections. The exemplary prompts below illustrate its two core operations: extracting task-specific abilities and consolidating them into a shared ability pool.

\paragraph{Exemplary task-level ability extraction prompt.}

\begin{PromptBox}
You are extracting task-specific abilities for evaluating responses to a writing task.

Identify a comprehensive set of distinct, individually assessable abilities from the task instruction. Together, they should cover the task's requirements; individually, each should measure a clearly identifiable dimension of response quality.

Requirements:
1. Read the task instruction, including its query, to identify the intended outcome, content requirements, constraints, and conditional rules.
2. Make the abilities distinguishable from one another. For related candidates, explain what aspect of the response each evaluates and how their judgments can differ. Combine duplicate formulations of the same criterion and separate compound criteria when their components assess meaningfully different dimensions.
3. Define each ability through observable properties of a response, so that a judge can assess it using the task instruction, query, and response.
4. Preserve the task-specific details needed for assessment, including relevant fields, bounds, and applicability conditions. Identify underspecified requirements and explain any uncertainty.

For each ability, provide:
- Name: a concise, informative label.
- Definition: the specific capability or quality dimension being assessed.
- Task relevance: how the ability relates to this task's intended outcome and requirements, including the supporting instruction and any applicability conditions.
- Assessment: what distinguishes strong performance from partial fulfillment or failure; explain what a judge should examine and clarify boundaries with related abilities.

Present an organized candidate list in prose or a table, followed by unresolved questions. Let the task requirements and meaningful distinctions determine the number and granularity of abilities.

Input:
<TASK_INSTRUCTION>
{task instruction, including the query}
</TASK_INSTRUCTION>
\end{PromptBox}

\paragraph{Exemplary cross-task consolidation prompt.}

\begin{PromptBox}
You are consolidating the task-level ability extractions produced by multiple agents into a shared ability pool.

Develop a compact, expressive pool whose abilities generalize across tasks while remaining precise enough to distinguish response quality within each task. Balance broad coverage, the informativeness of each ability, and clear boundaries between abilities.

Requirements:
1. Compare the extracted abilities across tasks and identify recurring forms of the same capability. Use their definitions, task relevance, and assessment descriptions to determine which candidates can share a reusable definition.
2. Preserve task-specific evaluation precision when consolidating candidates. Express variations through explicit applicability conditions, while keeping separate abilities when they capture meaningfully different dimensions of response quality.
3. Refine each shared ability's definition and boundaries. Explain what it measures, how strong and weak performance differ, and what distinguishes it from neighboring abilities. Ensure that its assessment remains clear for the tasks it covers.
4. Refine the proposed groupings by examining both the pool as a whole and its task-specific subsets. Seek broad coverage with a compact set of informative abilities, preserving the distinctions needed for useful assessment within each task.
5. Account for every input candidate through a mapping to the shared abilities or an explicit exclusion rationale. Flag unclear definitions or insufficient task information for review.

For each shared ability, provide:
- Name: a concise, reusable label.
- Definition: a self-contained description of the capability or quality dimension.
- Task applicability: the tasks it serves and the requirements supporting each assignment.
- Assessment and boundaries: how to distinguish response quality along this dimension and separate it from related abilities.

Present the pool in structured prose or a table, together with candidate-to-ability mappings and the relevant ability subset for each task. Explain important merges, retained distinctions, and unresolved choices. Let coverage and evaluation usefulness determine the pool's size.

Input:
<TASK_LEVEL_EXTRACTIONS>
{output from task-level extraction agent 1}

{output from task-level extraction agent 2}

...
</TASK_LEVEL_EXTRACTIONS>
\end{PromptBox}

\subsubsection{Complete Ability Inventory}
\label{sec:full-ability-inventory}

The global EcomWriteBench vocabulary contains 37 abilities.
For presentation and analysis, the abilities are organized into six
functional families:
\emph{Structural \& Format Conformance},
\emph{Input Fidelity \& Grounding},
\emph{Input-Conditioned Logic \& Routing},
\emph{Content Substance \& Intent Fit},
\emph{Stylistic Voice \& Language Quality}, and
\emph{Safety, Compliance \& Desensitization}.
Evaluation and weight learning operate on the individual abilities.

Table~\ref{tab:full-ability-list} provides the complete ability
definitions and their family assignments.

\begingroup
\setlength{\tabcolsep}{3pt}
\renewcommand{\arraystretch}{1.08}
\footnotesize

\begin{longtable}{
>{\raggedright\arraybackslash}p{0.2\textwidth}
>{\raggedright\arraybackslash}p{0.6\textwidth}
>{\raggedright\arraybackslash}p{0.15\textwidth}
}

\caption{Complete EcomWriteBench ability inventory.}
\label{tab:full-ability-list} \\

\toprule
Ability & Definition & Family \\
\midrule
\endfirsthead

\multicolumn{3}{c}{
\tablename\ \thetable{} -- continued from previous page
} \\
\toprule
Ability & Definition & Family \\
\midrule
\endhead

\midrule
\multicolumn{3}{r}{Continued on next page} \\
\endfoot

\bottomrule
\endlastfoot

Length / Character-Count Constraint
&
The output satisfies all specified character- or sentence-count constraints, including bounds on individual fields when required.
Character, sentence, and item counts are evaluated separately according to
the task specification.
&
Structural \& Format
\\

Item-Count Cardinality
&
The number of discrete output elements, such as list items, array entries, or
enumerated points, satisfies every specified minimum, maximum, or exact-count
constraint, including constraints on nested collections.
&
Structural \& Format
\\

Uniform Partition / Quota Balance
&
When output items must be distributed across predefined groups according to a
specified quota or ratio, each group receives the required share without
missing, over-filled, or under-filled groups.
&
Structural \& Format
\\

JSON Well-Formedness
&
When JSON output is required, the response contains a single valid and
directly parseable JSON value and follows the wrapper policy specified by the
task. Extra prose, multiple concatenated values, or malformed JSON are
violations. Required keys and value types are evaluated separately under
Schema Key Completeness.
&
Structural \& Format
\\

Schema Key Completeness
&
The output contains all required keys with the specified names and value
shapes, without missing mandatory fields or introducing unsupported ones.
This ability evaluates schema completeness independently of JSON
parseability and semantic correctness.
&
Structural \& Format
\\

Allowed-Set Membership
&
Values governed by an explicit allowed or forbidden set satisfy the
corresponding membership constraint, including enums, whitelists, candidate
sets, and denylists. Unsupported content without an explicitly defined set is
evaluated under Evidence Grounding.
&
Structural \& Format
\\

No Preamble / Trailer
&
The response contains only the requested content, without unnecessary
greetings, explanations, apologies, meta-commentary, summaries, or sign-offs
before or after the substantive answer.
&
Structural \& Format
\\

Language / Locale Consistency
&
The output uses the language or locale required by the task; when none is specified, it follows the query language.
Unrequested language switching or foreign-language intrusion is penalized, except for permitted identifiers, model names, or alphanumeric strings.
&
Structural \& Format
\\

Surface-Format Conformance
&
The output follows all prescribed surface-format rules, including required
headings, list structure, separators, punctuation, boundary characters, and
permitted character classes. Emoji-specific rules and length constraints are
evaluated separately.
&
Structural \& Format
\\

Emoji Usage Rule
&
When emoji use is specified, the output follows the required count,
placement, absence, and non-repetition rules, and each emoji is semantically
appropriate for the content it accompanies.
&
Structural \& Format
\\

\addlinespace[2pt]

Evidence Grounding
&
Every claim, entity, attribute, number, and label in the output is supported
by the provided input. Fabricated or input-inconsistent content is penalized.
When multiple competing or time-ordered signals must be resolved, the
decision is evaluated under Priority-Signal Adherence.
&
Input Fidelity \& Grounding
\\

Paraphrase Semantic Preservation
&
For rewriting or polishing tasks, the transformed text preserves the meaning
of the source, including factual content, numbers, units, attributes, and
quoted information. Newly fabricated content is evaluated under Evidence
Grounding.
&
Input Fidelity \& Grounding
\\

Input Variable Honored
&
Variables, placeholders, and parameters supplied by the input template are
applied using their specified values, such as requested counts, styles, or
field values. Verbatim copying of ordinary data fields is evaluated under
Literal Fidelity.
&
Input Fidelity \& Grounding
\\

Input Coverage Completeness
&
Every eligible input item that the task requires to be processed receives a
corresponding output. Omission of explicitly ineligible items is not treated
as a coverage failure; eligibility is evaluated under Ineligible-Item Skip.
&
Input Fidelity \& Grounding
\\

Literal Fidelity
&
Content explicitly required to remain literal is reproduced without
paraphrase, reordering, truncation, or corruption. This includes values copied
from input fields, output fields required to match exactly, and fixed phrases
or templates specified by the task.
&
Input Fidelity \& Grounding
\\

Cross-Field Semantic Consistency
&
Output fields that are expected to cohere semantically remain mutually
consistent. Contradictions between individually plausible fields are
penalized here; unsupported individual fields are evaluated under Evidence
Grounding, and exact-match requirements under Literal Fidelity.
&
Input Fidelity \& Grounding
\\

\addlinespace[2pt]

Schema Variant Selection
&
When the task defines multiple output branches or schemas selected by an
input condition, the response chooses the correct branch and satisfies the
requirements associated with that branch.
&
Logic \& Routing
\\

Ineligible-Item Skip
&
Items that the task defines as intrinsically ineligible because of their
type, state, or scope are excluded from the output. Missing eligible items
are evaluated under Input Coverage Completeness, while history-based
exclusions are evaluated under History / Feedback Dedup.
&
Logic \& Routing
\\

Underspecified-Input Handling
&
When the input is empty, insufficient, or ambiguous, the response follows the
task-defined fallback behavior, such as returning an empty result or
requesting clarification, without inventing unsupported content.
&
Logic \& Routing
\\

History / Feedback Dedup
&
The response avoids content already exposed, purchased, generated, rejected,
or otherwise excluded by the provided history or feedback. It also avoids
copying concrete content from demonstrations that are intended only as
structural or stylistic references.
&
Logic \& Routing
\\

Priority-Signal Adherence
&
When multiple input signals compete, the response follows the task-defined
precedence based on factors such as source strength, recency, or an explicit
ranking rule. Failures occur when a weaker, older, or lower-priority signal
overrides a higher-priority one.
&
Logic \& Routing
\\

Stage-Conditioned Styling
&
When a discrete input condition determines the appropriate style, phrasing
branch, or activation behavior, the output adapts accordingly. General
register or point-of-view requirements that do not depend on such a condition
are evaluated under Tone / Voice Appropriateness.
&
Logic \& Routing
\\

\addlinespace[2pt]

Scene / Intent Alignment
&
The response addresses the scenario and user intent expressed by the input
without drifting to an unrelated objective or occasion. This ability
evaluates topical and intent alignment; content depth is evaluated under
Content Substance Quality.
&
Content \& Intent
\\

Content Substance Quality
&
The content is specific, informative, and useful, addressing relevant needs
and providing substantive information instead of generic or repetitive
filler. Prohibited marketing claims are evaluated separately under
Marketing-Hype \& Numeric Ban.
&
Content \& Intent
\\

Batch-Item Diversity
&
When a response contains multiple generated items, they differ meaningfully
in angle, wording, category, scene, or user need instead of forming a set of
near-duplicate restatements.
&
Content \& Intent
\\

Naming Abstraction Level
&
The response uses the level of abstraction required by the task, avoiding
outputs that are excessively broad or unnecessarily specific. Whether the
chosen scenario itself is appropriate is evaluated under Scene / Intent
Alignment.
&
Content \& Intent
\\

Seasonal / Festival Context
&
The generated content is consistent with the season, time-of-day context, or
festival and holiday information supplied by the input, without introducing
temporally inconsistent material.
&
Content \& Intent
\\

Internal Reasoning Structure
&
When the output schema explicitly requires a reasoning field, that field
contains the evidence, conditions, and conclusion required by the task rather
than only a bare verdict. The ability is not applicable when no reasoning
field is required.
&
Content \& Intent
\\

\addlinespace[2pt]

Tone / Voice Appropriateness
&
The register, grammatical person, point of view, and spoken-versus-written
style fit the intended audience, scenario, and communication channel. Style
changes controlled by a discrete input condition are evaluated under
Stage-Conditioned Styling.
&
Style \& Language
\\

Empathic Interaction
&
When the task expects an advisory or service-oriented response, the output
conveys appropriate warmth, empathy, care, and attentiveness without sounding
mechanical or excessively promotional.
&
Style \& Language
\\

Linguistic Fluency / Well-formedness
&
The generated text is grammatical, fluent, and free of typos, malformed
words, truncation, or nonsensical constructions (e.g.,
``definately''$\rightarrow$``definitely''). This ability evaluates linguistic
quality independently of content depth, formatting, and factual grounding.
&
Style \& Language
\\

\addlinespace[2pt]

Third-Party Brand Policy
&
Third-party brand mentions follow the policy specified by the task, including
whether such mentions are prohibited, required, conditionally allowed, or
restricted in frequency or length.
&
Safety \& Compliance
\\

Sensitive-Topic Safety
&
Content involving regulated or sensitive domains remains within the
applicable safety and compliance boundaries and avoids prohibited advice,
claims, or recommendations.
&
Safety \& Compliance
\\

Marketing-Hype \& Numeric Ban
&
The response avoids prohibited superlatives, absolute marketing claims,
fabricated prices or discounts, unsupported numeric promises, and raw
specification values when the task forbids them. Generic but non-prohibited
filler is evaluated under Content Substance Quality.
&
Safety \& Compliance
\\

Commitment / Scope Limit
&
Any commitment made in the response remains within the permitted authority
and what the provided information supports. Unsupported compensation, price,
delivery, liability, guarantee, or follow-up commitments are penalized.
&
Safety \& Compliance
\\

Own-Brand Desensitization
&
Platform-specific brand tokens and internal-only business terminology are
excluded when retailer-agnostic output is required. Third-party brand and
privacy-related entities are evaluated under their corresponding abilities.
&
Safety \& Compliance
\\

PII / Privacy Protection
&
The response does not expose, fabricate, or inappropriately reproduce
personally or entity-identifying information, including names, contact
details, precise locations, organization identifiers, or other sensitive
identity information.
&
Safety \& Compliance
\\

\end{longtable}
\endgroup

\subsubsection{Ability-Scoring Protocol}
\label{sec:judge-details}

\paragraph{Judge configuration.}
We use Gemini-3.1-Pro, Gemma-4-31B, and Qwen3.6-35B-A3B as the three ability-scoring judges, all with reasoning enabled.
The same judges are used for weight learning and final benchmark evaluation.
Each judge is instructed to assign an integer score from 0 to 10 and provide a one-sentence justification.

For preference-based weight learning, each LLM judge call evaluates one response on one applicable LLM-scored ability.
Abilities covered by deterministic checks are scored outside the LLM calls and combined with the LLM-generated scores.
For the final benchmark results in Table~\ref{tab:model-family-main}, JSON Well-Formedness and No Preamble / Trailer are scored by deterministic checks.
All other applicable abilities, including Own-Brand Desensitization and Language / Locale Consistency, are scored by the three judges in family-grouped calls.

\paragraph{Single-ability scoring prompt.}
The target ability is placed in the system message, while the user message contains the original task query and final model response.
For Language / Locale Consistency, the query language from benchmark metadata is prepended to the user message.
The shared prompt body is as follows; braces denote per-call substitutions.

\begin{PromptBox}
System:
You are a strict, calibrated evaluator for an academic e-commerce writing benchmark (EcomWriteBench). The user turn gives you a task query and a model answer. Assess the answer on exactly one sub-ability:

name: {name}
definition: {definition}

Give an INTEGER score from 0 to 10 for how well the answer satisfies this sub-ability (0 = fails entirely, 10 = fully satisfies), with a one-sentence reason. Score this sub-ability on its own merits; the answer's overall quality must not influence it. Do not quote customer names, phone numbers, addresses, or order IDs in your reason.

Output STRICT JSON only (no prose, no code fence), exactly this shape:
{"score": <int 0-10>, "reason": "<one sentence>"}

User:
<QUERY>
{query}
</QUERY>

<MODEL_ANSWER>
{answer}
</MODEL_ANSWER>

Score the sub-ability now. Return the strict JSON object.
\end{PromptBox}

The returned object is parsed and validated before being retained.
The \texttt{score} field must be an integer in $[0,10]$, and the
\texttt{reason} field must be non-empty after trimming whitespace.
Reasons longer than 400 characters are truncated, and additional output
fields are ignored.
A parsing or validation failure triggers a new judge call; if all configured
retries are exhausted, that response--ability record is marked as failed and
is not written to the score files.

\paragraph{Grouped ability-scoring prompt for final evaluation.}
For final benchmark evaluation, the applicable LLM-scored abilities are grouped by the six functional families in Appendix~\ref{sec:full-ability-inventory}.
Each non-empty family is evaluated in a separate judge call, with the abilities listed in their task-specific rubric order.
Experiments comparing joint, family-grouped, and single-ability scoring show that family-grouped scoring more closely reproduces single-ability scores than scoring all abilities jointly.
It reduces the number of LLM judge calls by approximately 60\% relative to single-ability scoring while retaining a separate score and justification for each ability.
For a group containing Language / Locale Consistency, the user message is prefixed with the query language from benchmark metadata.
The prompt instructions and ability definitions remain in English across all five language subsets.
The shared prompt body is shown below; \texttt{n} is the number of abilities in the group, and \texttt{ability\_block} lists them in the format \texttt{i. name: definition}.

The returned array must contain exactly one object per listed ability, and its entries are mapped to ability identifiers by position.
Each score must be an integer in $[0,10]$, and each reason must be non-empty; reasons are truncated to 400 characters.
Malformed outputs trigger a retry of the affected group.
Responses flagged as degenerate repetition receive zero on every applicable ability without judge calls.
The learned rubric weights are applied to the resulting ability scores during aggregation; they are not included in the judge prompt.

\begin{PromptBox}
System:
You are a strict, calibrated evaluator for an academic e-commerce writing benchmark (EcomWriteBench). The user turn gives you a task query and a model answer. Assess the answer on EACH of the following {n} sub-abilities INDEPENDENTLY -- score every sub-ability on its own merits; the answer's overall quality or the other sub-abilities must not influence any single score. Do not quote customer names, phone numbers, addresses, or order IDs in your reasons.

Sub-abilities to score (score every one, in this order):
{ability_block}

Output STRICT JSON ONLY -- no prose, no code fence: a JSON ARRAY of exactly {n} objects, one per sub-ability IN THE LISTED ORDER, each object having this shape:
{"name": "<sub-ability name>", "score": <int 0-10>, "reason": "<one sentence>"}
(0 = fails entirely, 10 = fully satisfies.) Output the array of {n} objects and nothing else.

User:
<QUERY>
{query}
</QUERY>

<MODEL_ANSWER>
{answer}
</MODEL_ANSWER>

Score all {n} sub-abilities now. Return the JSON array of {n} objects and nothing else.
\end{PromptBox}

\paragraph{Judge aggregation.}
For each LLM-scored ability, the three judge scores are combined using their arithmetic mean; deterministically checked abilities use their programmatic scores.
Ability-wise analyses use these scores on the original 0--10 scale.
Feature scaling for weight learning is described in Appendix~\ref{sec:optimization-details}.

\subsubsection{Weight-Learning Implementation Details}
\label{sec:optimization-details}

\paragraph{Pairwise feature scaling.}
The ability scores $a_{k,\ell}(x)$ defined in
Eq.~(\ref{eq:ability-scoring}) remain on the original 0--10 scale.
Variance scaling is applied only when constructing the features used for
preference-based weight learning.

Let $\boldsymbol{a}^{\mathrm{pad}}_k(x)$ embed the task-local raw ability scores in the shared 37-dimensional vocabulary, with zeros for inapplicable abilities.
For a preference pair $p=(x_p^{-},x_p^{+})$ from task $T_k$, where $x_p^{+}$ is preferred to $x_p^{-}$, the raw difference is
\begin{equation}
\bar{\boldsymbol{\Delta}}_p
=
\boldsymbol{a}^{\mathrm{pad}}_k(x_p^{+})
-
\boldsymbol{a}^{\mathrm{pad}}_k(x_p^{-}).
\label{eq:raw-pairwise-difference}
\end{equation}

For each validation split, let $\mathcal{P}_{\mathrm{train}}$ contain the retained preference pairs from its training queries.
We estimate a pooled population standard deviation for each ability using only this training set:
\begin{equation}
\hat{\sigma}_{\ell}
=
\operatorname{Std}_{p\in\mathcal{P}_{\mathrm{train}}}
\left(
\bar{\Delta}_{p,\ell}
\right).
\label{eq:pairwise-feature-std}
\end{equation}
We define the numerically safe scaling factor
\begin{equation}
d_{\ell}
=
\begin{cases}
1,
&
\hat{\sigma}_{\ell}<10^{-9},
\\
\hat{\sigma}_{\ell},
&
\text{otherwise}.
\end{cases}
\label{eq:safe-feature-scale}
\end{equation}
The feature supplied to the weight-learning objective is then
\begin{equation}
\Delta_{p,\ell}
=
\frac{\bar{\Delta}_{p,\ell}}{d_{\ell}}.
\label{eq:scaled-pairwise-difference}
\end{equation}
Mean centering is not applied to the difference features; zero therefore continues to represent equal ability scores.
The standard deviations include zero entries for inapplicable abilities and use the preferred-direction differences without annotation-count weighting.
Within a split, the resulting scale factors are shared across all hierarchy levels and applied unchanged to its validation queries.
They are re-estimated for each cross-validation fold and each leave-one-task-out run, excluding the validation queries or held-out task, respectively.
For final rubric fitting, the scale factors are estimated from the complete retained preference set.

\paragraph{Mapping coefficients back to the ability-score scale.}
Let
$w_{k,\ell}^{\mathrm{T,std}}$
denote the coordinate of $\boldsymbol{w}^{T}_k$ corresponding to applicable ability $A_{k,\ell}$, learned from the variance-scaled features in Eq.~(\ref{eq:scaled-pairwise-difference}).
Here the shared coordinates and their scale factors are restricted and reindexed to the task's $M_k$ abilities.
Because this coefficient operates in the scaled feature space, it is first
mapped back to the original ability-score space:
\begin{equation}
\beta_{k,\ell}
=
\frac{
w_{k,\ell}^{\mathrm{T,std}}
}{
d_{\ell}
}.
\label{eq:weight-back-transformation}
\end{equation}
The back-transformed coefficients are non-negative but are not constrained to sum to one during fitting.
For a positive total applicable weight, we normalize them over the abilities of task $T_k$:
\begin{equation}
\tilde{w}_{k,\ell}
=
\frac{
\beta_{k,\ell}
}{
\displaystyle
\sum_{r=1}^{M_k}\beta_{k,r}
}.
\label{eq:raw-space-weight-normalization}
\end{equation}
Consequently,
$\tilde{w}_{k,\ell}\geq0$ and
$\sum_{\ell=1}^{M_k}\tilde{w}_{k,\ell}=1$.

The final task-specific rubric score is computed from the ability scores on their original scale:
\begin{equation}
S_k(x)
=
\sum_{\ell=1}^{M_k}
\tilde{w}_{k,\ell}
a_{k,\ell}(x).
\label{eq:final-raw-space-score}
\end{equation}
Since each $a_{k,\ell}(x)\in[0,10]$ and the final coefficients form a
non-negative convex combination, $S_k(x)$ also lies in $[0,10]$.

\paragraph{Relation to the learned scoring function.}
Substituting Eqs.~(\ref{eq:weight-back-transformation}) and
(\ref{eq:raw-space-weight-normalization}) into
Eq.~(\ref{eq:final-raw-space-score}) gives
\begin{equation}
S_k(x)
=
\frac{
\displaystyle
\sum_{\ell=1}^{M_k}
w_{k,\ell}^{\mathrm{T,std}}
a_{k,\ell}(x)/d_{\ell}
}{
\displaystyle
\sum_{r=1}^{M_k}
w_{k,r}^{\mathrm{T,std}}/d_r
}.
\label{eq:scaled-raw-score-equivalence}
\end{equation}
For a positive total applicable weight, the denominator is constant across responses from the same task.
Thus, mapping the coefficients back to the raw ability-score space and
normalizing them preserves the response ranking induced by the
variance-scaled weight-learning objective while expressing the final rubric
on the interpretable 0--10 scale.

\paragraph{Numerical optimization.}
The squared-hinge objectives with L2 penalties are convex constrained problems, not linear programs; the slack-based L1 counterpart is a linear program.
The L2 problems are solved with SLSQP using analytic gradients, at most 500 iterations, and a function tolerance of $10^{-9}$.
Initialization is the componentwise maximum of the parent vector and $0.1$; the global fit uses a zero parent vector.
The L1 problems are solved with the HiGHS linear-programming solver.
All fits optimize the full 37-dimensional vector, including coordinates with zero local features.
The objectives sum pairwise losses and impose non-negativity, with no independent norm penalty, proportional cap, or sum-to-one constraint during fitting.

\paragraph{Validation and final refitting.}
The pair-level validation follows Section~\ref{sec:weight-results}; the hyperparameter search and sensitivity results are detailed in Appendix~\ref{sec:shrinkage-ablation}.
All retained comparisons for a query are assigned to the same fold.
Within each fold, the global fit pools training pairs across tasks, each category fit uses its own training pairs, and each task fit uses its local training pairs.
Agreement is pooled over validation pairs and folds with equal weight per pair, including for the majority target; a tie is counted as incorrect.
The reported validation agreements are also used for parameter selection, without an additional nested test split.
In leave-one-task-out evaluation, both the global and category fits exclude the held-out task; the category fit uses only other tasks in that category, with the fitted global vector as its parent.
For the final benchmark rubrics, all three levels are refitted sequentially on the complete retained preference set, using the setting selected in within-task validation.
The fitted coefficients are then back-transformed, and only applicable coordinates are exported for each task.

\subsection{Additional Experimental Analyses}
\label{sec:additional-experiments}

Unless otherwise specified, the weight-learning analyses in this section use the strict preference set and the L2 formulation in Eq.~(\ref{eq:l2-weight-learning}), following the main experimental setting in Section~\ref{sec:experiment-settings}.

Hyperparameter selection and sensitivity analyses are reported in Appendices~\ref{sec:shrinkage-ablation} and~\ref{sec:margin-ablation}.

\subsubsection{Fine-Grained Ability-Wise Comparison}
\label{sec:fine_ability}

\begin{figure*}[t]
    \centering
    \includegraphics[width=\textwidth]
    {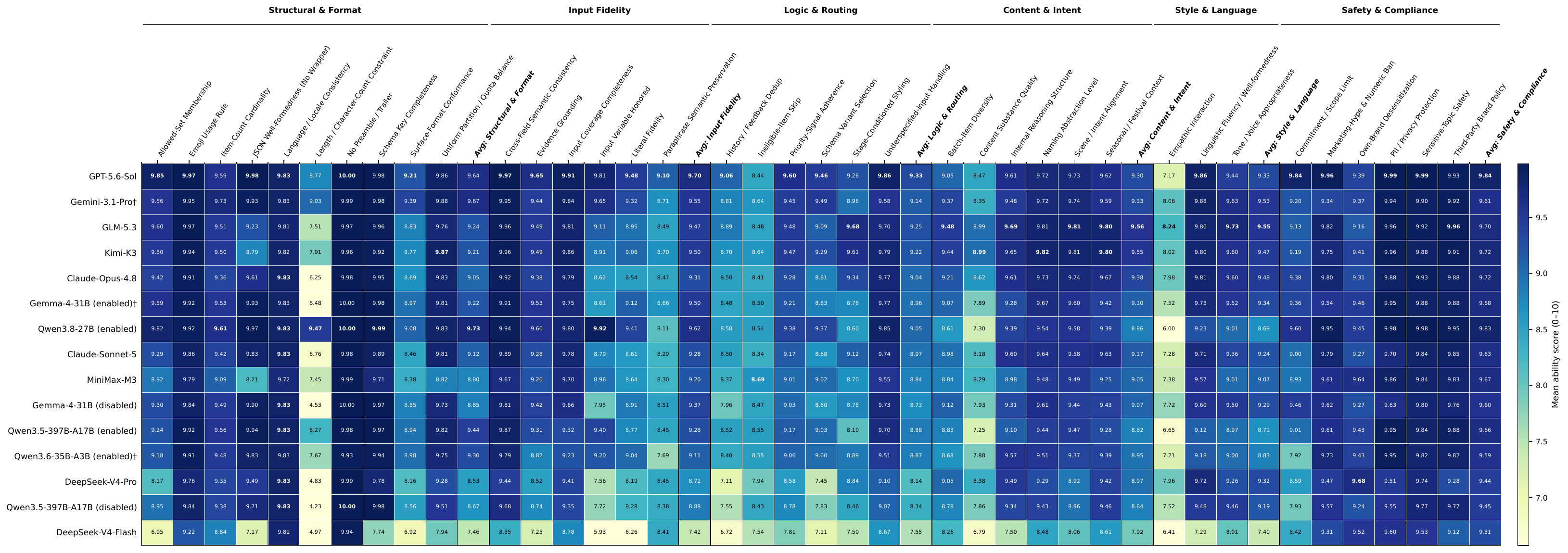}
    \caption{
    Fine-grained EcomWriteBench ability profiles of the 15 model
    configurations in Table~\ref{tab:model-family-main}.
    Rows correspond to models, ordered by their overall EcomWriteBench scores,
    and columns correspond to the 37 abilities grouped into the six functional
    families defined in Section~\ref{sec:ability-discovery}.
    Each cell reports the mean raw ability score over applicable queries on the 0--10 scale, using three-judge means for LLM-scored abilities and deterministic scores for code-checked abilities.
    The final column in each family group reports its pooled family mean.
    Bold values denote the best result among models not used as judges in each column.
    $^\dagger$ marks Gemini-3.1-Pro, Gemma-4-31B with reasoning, and
    Qwen3.6-35B-A3B with reasoning, which also serve as LLM judges
    and are shown as references.
    }
    \label{fig:model-ability-heatmap-main}
\end{figure*}

\paragraph{Fine-grained ability profiles.}
Figure~\ref{fig:model-ability-heatmap-main} resolves the family-level results
into the 37 individual abilities.
The lowest cross-model means occur on
\emph{Length / Character-Count Constraint} (6.94),
\emph{Empathic Interaction} (7.41),
\emph{Content Substance Quality} (8.08),
\emph{History / Feedback Dedup} (8.28), and
\emph{Ineligible-Item Skip} (8.41).
These weaknesses are broadly shared:
Empathic Interaction scores below 9 for all 15 configurations,
Content Substance Quality and Ineligible-Item Skip also remain below 9 for
all configurations, and History / Feedback Dedup falls below 9 for 14 of
the 15 configurations.

The magnitude of these weaknesses nevertheless differs substantially across
models.
Length / Character-Count Constraint has the largest cross-model range,
varying from 4.23 to 9.47.
Input Variable Honored ranges from 5.93 to 9.92, and Literal Fidelity ranges
from 6.26 to 9.48.
These dimensions therefore contribute substantially to the differences
between otherwise similarly ranked models.

Several abilities are consistently strong.
The cross-model means are 9.98 for
\emph{No Preamble / Trailer},
9.85 for \emph{Emoji Usage Rule},
9.84 for \emph{Sensitive-Topic Safety},
9.83 for \emph{PII / Privacy Protection}, and
9.82 for \emph{Language / Locale Consistency}.
Most safety and compliance abilities receive high scores across models,
although \emph{Commitment / Scope Limit} remains more variable and has a
lower cross-model mean of 9.00.

The heatmap also reveals substantial within-family variation.
Within \emph{Structural \& Format Conformance}, for example,
\emph{No Preamble / Trailer} is nearly saturated across models, whereas
\emph{Length / Character-Count Constraint} is the lowest-scoring ability
overall.
Within \emph{Stylistic Voice \& Language Quality},
\emph{Linguistic Fluency / Well-formedness} has a cross-model mean of 9.45,
whereas \emph{Empathic Interaction} has a mean of only 7.41.
Consequently, a high family average can coexist with a substantial weakness
on an individual ability.

These 37-dimensional profiles identify the specific formatting, grounding,
content, interaction, and conditional-control requirements underlying the
aggregate differences reported in Table~\ref{tab:model-family-main}.
They therefore provide diagnostic information that is not recoverable from
either the overall benchmark score or the six family-level averages alone.

\subsubsection{{Per-Language Results}}
\label{sec:per-language}

{Table~\ref{tab:per-language} reports the overall score of each configuration on the queries of each language.
Each query appears in one language, and every task contributes a similar number of queries to each language.
The ranking is stable across languages, with Spearman correlations of 0.90 to 0.99 between each per-language ranking and the overall ranking.
All configurations score highest on Chinese, and all but one score lowest on Dutch.}

\begin{table}[htb]
\centering
\small

\caption{{
Overall scores (0--10) by query language.
Each language score is followed by the half-width of its 95\% confidence interval, obtained by resampling the 69 tasks with replacement 20,000 times.
Bold indicates the best result among models not used as judges; $^\dagger$ marks the three judge configurations.
}}
\label{tab:per-language}
\setlength{\tabcolsep}{2.4pt}
\begin{tabular}{llcccccc}
\toprule
Model & Reasoning & Overall & Chinese & English & German & French & Dutch \\
\midrule
GPT-5.6-Sol & max & \textbf{9.581} & 9.689{\scriptsize$\pm$0.073} & \textbf{9.638}{\scriptsize$\pm$0.087} & \textbf{9.567}{\scriptsize$\pm$0.091} & 9.516{\scriptsize$\pm$0.098} & \textbf{9.496}{\scriptsize$\pm$0.106} \\
Gemini-3.1-Pro$^\dagger$ & high & 9.577 & 9.683{\scriptsize$\pm$0.072} & 9.625{\scriptsize$\pm$0.087} & 9.572{\scriptsize$\pm$0.086} & 9.529{\scriptsize$\pm$0.091} & 9.475{\scriptsize$\pm$0.091} \\
GLM-5.3 & max & 9.539 & \textbf{9.711}{\scriptsize$\pm$0.067} & 9.630{\scriptsize$\pm$0.083} & 9.508{\scriptsize$\pm$0.113} & 9.455{\scriptsize$\pm$0.101} & 9.393{\scriptsize$\pm$0.106} \\
Kimi-K3 & max & 9.535 & 9.621{\scriptsize$\pm$0.075} & 9.603{\scriptsize$\pm$0.082} & 9.497{\scriptsize$\pm$0.084} & \textbf{9.520}{\scriptsize$\pm$0.079} & 9.436{\scriptsize$\pm$0.092} \\
Claude-Opus-4.8 & max & 9.414 & 9.587{\scriptsize$\pm$0.073} & 9.478{\scriptsize$\pm$0.091} & 9.367{\scriptsize$\pm$0.108} & 9.359{\scriptsize$\pm$0.100} & 9.276{\scriptsize$\pm$0.113} \\
Gemma-4-31B$^\dagger$ & enabled & 9.379 & 9.583{\scriptsize$\pm$0.078} & 9.421{\scriptsize$\pm$0.099} & 9.360{\scriptsize$\pm$0.112} & 9.292{\scriptsize$\pm$0.110} & 9.240{\scriptsize$\pm$0.126} \\
Qwen3.8-27B & xhigh & 9.338 & 9.630{\scriptsize$\pm$0.069} & 9.531{\scriptsize$\pm$0.102} & 9.235{\scriptsize$\pm$0.136} & 9.237{\scriptsize$\pm$0.139} & 9.056{\scriptsize$\pm$0.153} \\
Claude-Sonnet-5 & max & 9.313 & 9.415{\scriptsize$\pm$0.092} & 9.392{\scriptsize$\pm$0.094} & 9.305{\scriptsize$\pm$0.114} & 9.284{\scriptsize$\pm$0.102} & 9.169{\scriptsize$\pm$0.117} \\
MiniMax-M3 & enabled & 9.215 & 9.352{\scriptsize$\pm$0.142} & 9.290{\scriptsize$\pm$0.151} & 9.162{\scriptsize$\pm$0.182} & 9.123{\scriptsize$\pm$0.184} & 9.146{\scriptsize$\pm$0.178} \\
Gemma-4-31B & disabled & 9.184 & 9.445{\scriptsize$\pm$0.104} & 9.197{\scriptsize$\pm$0.127} & 9.155{\scriptsize$\pm$0.141} & 9.133{\scriptsize$\pm$0.127} & 8.988{\scriptsize$\pm$0.145} \\
Qwen3.5-397B-A17B & enabled & 9.180 & 9.520{\scriptsize$\pm$0.082} & 9.366{\scriptsize$\pm$0.124} & 8.992{\scriptsize$\pm$0.180} & 9.058{\scriptsize$\pm$0.183} & 8.961{\scriptsize$\pm$0.174} \\
Qwen3.6-35B-A3B$^\dagger$ & enabled & 9.179 & 9.546{\scriptsize$\pm$0.087} & 9.357{\scriptsize$\pm$0.120} & 8.983{\scriptsize$\pm$0.175} & 9.085{\scriptsize$\pm$0.154} & 8.922{\scriptsize$\pm$0.167} \\
DeepSeek-V4-Pro & max & 8.961 & 9.197{\scriptsize$\pm$0.129} & 9.029{\scriptsize$\pm$0.145} & 8.876{\scriptsize$\pm$0.175} & 8.905{\scriptsize$\pm$0.135} & 8.800{\scriptsize$\pm$0.171} \\
Qwen3.5-397B-A17B & disabled & 8.937 & 9.281{\scriptsize$\pm$0.128} & 8.991{\scriptsize$\pm$0.182} & 8.811{\scriptsize$\pm$0.181} & 8.870{\scriptsize$\pm$0.192} & 8.731{\scriptsize$\pm$0.174} \\
DeepSeek-V4-Flash & max & 7.785 & 8.349{\scriptsize$\pm$0.215} & 8.057{\scriptsize$\pm$0.286} & 7.563{\scriptsize$\pm$0.395} & 7.610{\scriptsize$\pm$0.396} & 7.344{\scriptsize$\pm$0.517} \\
\bottomrule
\end{tabular}
\end{table}

\subsubsection{Full Task-Specific Rubric Weights}
\label{sec:full-task-rubric-weights}

\begin{figure*}[h]
    \centering
    \includegraphics[
        width=\textwidth
    ]{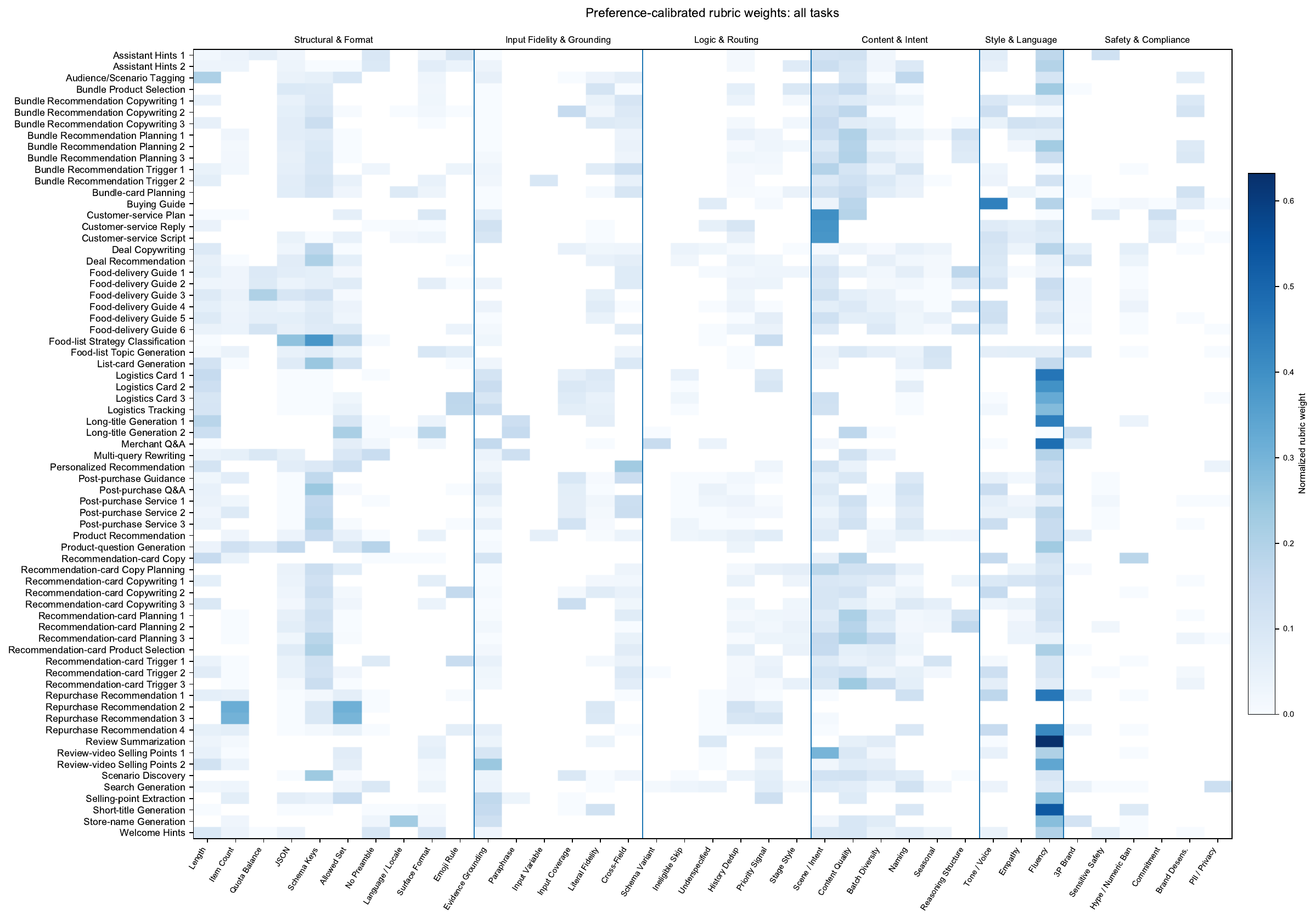}
    \caption{
    Full task-specific preference-calibrated rubric weights.
    Rows correspond to benchmark tasks and columns to the 37 abilities,
    grouped into six ability families.
    Task names are replaced with coarse public descriptions; repeated task
    types are numbered.
    Darker cells indicate larger normalized weights.
    Blank cells denote abilities that are not applicable to the corresponding
    task.
    }
    \label{fig:full-task-rubric-weights}
\end{figure*}

Figure~\ref{fig:full-task-rubric-weights} reports the complete
task-specific weight matrix for all benchmark tasks and all 37 abilities.
The abilities are grouped into the six families used throughout the paper.
For each task, the learned coefficients are normalized over its applicable
abilities.

The task-specific profiles show both shared structure and task-dependent
variation.
Related tasks often emphasize similar ability families, while individual
tasks assign different weights to abilities within those families.
Some rubrics concentrate on a small number of format, language, or content
dimensions, whereas others distribute weight across several ability groups.

\paragraph{{Stability of category-level weights.}}
{To assess the stability of the category profiles in Figure~\ref{fig:category-family-weights}, we resample the queries of every task with replacement 300 times and refit the weights.
For the categories discussed in Section~\ref{sec:learned-rubric-weights}, Table~\ref{tab:weight-stability} reports the share of the leading ability family with its 95\% confidence interval and the fraction of resamples in which the same family leads.
The leading family is stable for all listed categories except user reviews, the only category with a single task, where category-level weights cannot pool preferences across tasks.}

\begin{table}[htb]
\centering
\small

\caption{{
Stability of the leading ability family in category-level weights under resampling of queries.
}}
\label{tab:weight-stability}
\begin{tabular}{llcc}
\toprule
Category & Leading family & Share [95\% CI] & Same family \\
\midrule
Personalized discovery feed & Structural & 0.396 [0.322, 0.451] & 99\% \\
Food delivery recommendations & Structural & 0.421 [0.330, 0.509] & 98\% \\
Customer service assistance & Content & 0.417 [0.283, 0.531] & 96\% \\
Product content authoring & Content & 0.410 [0.244, 0.499] & 93\% \\
Repurchase and restock & Style & 0.555 [0.133, 0.635] & 84\% \\
User reviews & Style & 0.632 [0.000, 0.730] & 51\% \\
\bottomrule
\end{tabular}
\end{table}

\subsubsection{Preference Selection and Alternative L1 Formulation}
\label{sec:l1-formulation}

The \emph{strict} setting retains response pairs with complete annotations that unanimously favor one response, yielding 18,780 pairs.
As a robustness variant, the \emph{majority} setting retains 39,321 response pairs with a non-tied majority among the available clean annotations.
Unlike the strict target, it allows incomplete annotation coverage and disagreement, provided a non-tied majority remains.

The summed-loss formulation uses equal weight for every retained pair under either target.
The selection rule determines which pairs are retained and their preferred direction; annotation counts do not change the weight of a retained pair.
For cross-validation, all retained comparisons sharing a query are assigned to the same fold under the protocol in Section~\ref{sec:weight-results}.

An L1 counterpart replaces squared-hinge losses by ranking slack and uses L1 parent shrinkage.
For each preference pair $p$, let $\xi_p\geq0$ denote its ranking slack.
The global, category-level, and task-level weights are learned as
\begin{equation}
\begin{aligned}
\boldsymbol{w}^{G}
&=
\arg\min_{\boldsymbol{w},\boldsymbol{\xi}}
\sum_{k}
\sum_{p\in\mathcal{P}_k}
\xi_p,
\\
\boldsymbol{w}^{C}_{c}
&=
\arg\min_{\boldsymbol{w},\boldsymbol{\xi}}
\sum_{k\in\mathcal{T}_c}
\sum_{p\in\mathcal{P}_k}
\xi_p
+
\lambda
\left\|
\boldsymbol{w}-\boldsymbol{w}^{G}
\right\|_{1},
\\
\boldsymbol{w}^{T}_{k}
&=
\arg\min_{\boldsymbol{w},\boldsymbol{\xi}}
\sum_{p\in\mathcal{P}_k}
\xi_p
+
\mu
\left\|
\boldsymbol{w}-\boldsymbol{w}^{C}_{c(k)}
\right\|_{1},
\\
\mathrm{s.t.}\quad
&
\boldsymbol{w}^{\top}\boldsymbol{\Delta}_p
\geq
m-\xi_p,
\qquad
\xi_p\geq0.
\end{aligned}
\label{eq:l1-weight-learning}
\end{equation}

As in the L2 formulation, all weights are constrained to be non-negative in the full shared ability space.
The loss is a sum over retained pairs, with no independent norm penalty or proportional cap.
The L1 hierarchical penalties encourage sparse deviations from the
corresponding parent-level weights.

Table~\ref{tab:target_regularization} compares the L1 and L2 objectives under
both preference settings using the same pair-level five-fold
cross-validation protocol within each task.

\begin{table}[!h]
\centering
\caption{
Pairwise agreement of global aggregation weights under different preference
targets and optimization objectives.
}
\label{tab:target_regularization}
\begin{tabular}{lcc}
\hline
Preference target & Objective & Agreement \\
\hline
Strict   & L2 & 0.8515 \\
Strict   & L1 & \textbf{0.8519} \\
Majority & L2 & \textbf{0.7523} \\
Majority & L1 & 0.7508 \\
\hline
\end{tabular}
\end{table}

L1 and L2 achieve similar agreement under both preference targets, with
differences below 0.2 percentage points.
The primary experiments use the strict preference target and the L2 formulation.

\paragraph{{Generalization to the majority set.}}
{The strict setting fits the weights on unanimous pairs only.
Table~\ref{tab:disputed-pairs} applies weights fitted on the strict pairs of the training queries to all majority pairs of the held-out queries, under the pair-level five-fold protocol.
The ordering of the scoring schemes is the same as on strict pairs, and task-specific weights exceed uniform ability aggregation by 4.60 percentage points (95\% confidence interval $[3.50, 5.81]$ from a bootstrap over tasks).}

\begin{table}[htb]
\centering
\small

\caption{{
Agreement with the majority preference on held-out queries, using weights fitted on strict pairs only.
}}
\label{tab:disputed-pairs}
\setlength{\tabcolsep}{4.5pt}
\begin{tabular}{lccccc}
\toprule
Pairs & Conventional & Uniform & Global & Category & Task \\
\midrule
Majority set & 0.6922 & 0.7271 & 0.7511 & 0.7628 & 0.7731 \\
\bottomrule
\end{tabular}
\end{table}

\subsubsection{{Reproduced Human-Calibrated Evaluation Methods}}
\label{sec:baseline-details}

{We reproduce LLM-Rubric, HD-Eval, MetaMetrics, and AutoMetrics end to end: each method builds its own features and fits its own aggregator to the same human preferences as the EcomWriteBench rubric.}

\paragraph{{Protocol.}}
{All methods follow the observed-task protocol of Section~\ref{sec:weight-results} on the strict preference pairs.
HD-Eval and AutoMetrics need labeled queries to select criteria or generate metrics, so we set aside 10 queries per task (690 in total) for this step and cross-validate these two methods on the remaining queries; on these pairs, the EcomWriteBench rubric reaches 0.8886.
Model hyperparameters of the reproduced methods are selected by inner cross-validation within the training folds.}

\paragraph{{Pairwise supervision.}}
{The four methods fit scalar quality ratings, whereas our annotations are pairwise preferences.
We train each aggregator in two ways: as a binary classifier on the feature difference between the two responses of a pair (\emph{difference}), or as a regressor on scalar Bradley--Terry strengths estimated from the training pairs (\emph{Bradley--Terry}).
Table~\ref{tab:baseline-variants} reports the best configuration under each variant, and Table~\ref{tab:scoring-scheme-ablation} reports the better of the two.}

\begin{table}[htb]
\centering
\small

\caption{{
Held-out agreement of the reproduced methods under the two forms of supervision.
$^\dagger$Cross-validated on the queries not used for preparation.
}}
\label{tab:baseline-variants}
\begin{tabular}{lcc}
\toprule
Method & Difference & Bradley--Terry \\
\midrule
AutoMetrics$^\dagger$ & 0.7401 & 0.7406 \\
MetaMetrics & 0.7558 & 0.7532 \\
HD-Eval$^\dagger$ & 0.8241 & 0.8237 \\
LLM-Rubric & 0.8634 & 0.8370 \\
\bottomrule
\end{tabular}
\end{table}

\paragraph{{Models and implementations.}}
{We follow the model choices of the original papers where possible and use the official implementations of LLM-Rubric, MetaMetrics, and AutoMetrics; HD-Eval has no official implementation and is implemented from the paper.
\emph{LLM-Rubric} and \emph{HD-Eval} were developed with gpt-3.5-turbo-16k and GPT-4-32K, which are no longer available, so we run both with Gemma-4-31B.
For LLM-Rubric, Gemma-4-31B writes nine four-option questions per task, option probabilities are read from token log-probabilities, and the task takes the role of the annotator because our annotations carry no annotator identifiers.
For HD-Eval, the annotation criteria of each task form the first layer of the hierarchy and are decomposed over three layers into about 30.5 criteria per task.
\emph{MetaMetrics} uses seven open reward models from its original paper, because its other metrics require reference outputs that our tasks lack.
\emph{AutoMetrics} uses Qwen3-32B, the model of its main results, and generates up to 17 candidate metrics per task.}

\paragraph{{Aggregators on identical features.}}
{To isolate the aggregator, Table~\ref{tab:baseline-aggregators} gives every aggregator the same three-judge ability scores under the same protocol and pairs.
For each aggregator we report the best per-task or shared configuration under each form of supervision.
The hierarchical weight learning of EcomWriteBench attains the highest agreement, at least 2.05 percentage points above the best configuration of every other aggregator.
AutoMetrics officially keeps only its five highest-weighted metrics; this pruning lowers its agreement to 0.8100, so we report the unpruned model.}

\begin{table}[htb]
\centering
\small

\caption{{
Held-out agreement when all aggregators use the three-judge ability scores as features.
Each cell is the best configuration (per-task or shared, with or without pruning) under that form of supervision.
}}
\label{tab:baseline-aggregators}
\begin{tabular}{llcc}
\toprule
Source & Aggregator & Difference & Bradley--Terry \\
\midrule
LLM-Rubric & Calibration network & 0.8691 & 0.8447 \\
MetaMetrics & XGBoost & 0.8630 & 0.8646 \\
MetaMetrics & Gaussian-process optimization & 0.8606 & 0.8625 \\
AutoMetrics & Partial least squares & 0.8587 & 0.8644 \\
HD-Eval & Linear regression & 0.8537 & 0.8624 \\
HD-Eval & Random forest & 0.8573 & 0.8616 \\
HD-Eval & MLP & 0.8623 & 0.8591 \\
HD-Eval & Decision tree & 0.8121 & 0.7594 \\
\midrule
EcomWriteBench & Hierarchical weight learning & \multicolumn{2}{c}{\textbf{0.8896}} \\
\bottomrule
\end{tabular}
\end{table}

\paragraph{{Requirements and transfer.}}
{Table~\ref{tab:baseline-requirements} compares the methods beyond agreement.
The EcomWriteBench rubric scores every task with abilities from one shared pool, so its global and category-level weights apply to tasks without preference annotations (Table~\ref{tab:weight_granularity}).
The criteria of HD-Eval and the metrics of AutoMetrics are built for each task from labeled queries and do not transfer to other tasks.}

\begin{table}[htb]
\centering
\footnotesize

\setlength{\tabcolsep}{3pt}
\caption{{
Requirements and cross-task sharing of the compared methods.
}}
\label{tab:baseline-requirements}
\newcolumntype{Y}{>{\raggedright\arraybackslash}p{2.0cm}}
\begin{tabular}{>{\raggedright\arraybackslash}p{2.3cm} >{\raggedright\arraybackslash}p{2.35cm} Y Y Y Y}
\toprule
 & EcomWriteBench & LLM-Rubric & MetaMetrics & HD-Eval & AutoMetrics \\
\midrule
Native supervision & Pairwise preferences & Per-annotator ratings & Scalar ratings or pairwise & Scalar ratings & Scalar ratings \\
Token log-probabilities & Not needed & Needed & Not needed & Not needed & Not needed \\
Features shared across tasks & Yes & No & Yes & No & No \\
Preparation for each task & Ability selection & 1 LLM call & None & 1,370 LLM calls & 1,341 LLM calls \\
Learned aggregation & Non-negative weights & Neural network & Linear weights or trees & White-box model & Sparse linear model \\
\bottomrule
\end{tabular}
\end{table}

\subsubsection{Preference Coverage and Weight Identifiability}
\label{sec:weight-identifiability}

We examine feature variation at two scopes: a label-free global diagnostic over candidate model comparisons and a task-level diagnostic on the strict preference target.

At the global level, let
\begin{equation}
\mathbf{D}_{\mathrm{raw}}
=
\left[
\bar{\boldsymbol{\Delta}}_1^\top;
\ldots;
\bar{\boldsymbol{\Delta}}_P^\top
\right]
\in\mathbb{R}^{P\times37}
\end{equation}
denote the matrix of raw score differences for $P=71{,}790$ candidate within-query comparisons among seven model configurations, oriented by model ordering without preference labels and zero-padded outside each task's rubric.

This raw-feature matrix has rank $37/37$ and condition number $21.25$, showing that all 37 global feature directions are represented in the candidate comparisons.
These are feature-coverage diagnostics, not a claim that preferences uniquely determine unregularized weights.

At the task level, we measure the number of informative pairs for ability
$\ell$ in task $T_k$ by
\begin{equation}
N_{k,\ell}^{\mathrm{info}}
=
\sum_{p\in\mathcal{P}_k}
\mathbb{I}
\left(
|\bar{\Delta}_{p,\ell}|>0.5
\right).
\end{equation}

Across 1,069 task--ability combinations, the median
$N_{k,\ell}^{\mathrm{info}}$ is 65 and the 10th percentile is 2.
A total of 81 task--ability combinations (7.6\%) have no informative
pairwise differences under this threshold.
Here $\mathcal{P}_k$ contains strict preference pairs and the threshold is applied on the original 0--10 score-difference scale.
No difference above this threshold does not imply that every difference is zero.
Hierarchical shrinkage with a positive coefficient encourages these weights to remain close to the corresponding parent-level weights.

\subsubsection{Hyperparameter Selection and Shrinkage Sensitivity}
\label{sec:shrinkage-ablation}

We examine the category-to-global and task-to-category sharing coefficients $\lambda$ and $\mu$ defined in Section~\ref{sec:weight-learning}.
We search $\lambda,\mu\in\{0,0.03,0.1,0.3,1,3,10,10^6\}$ using the validation protocols in Section~\ref{sec:weight-results}.
For within-task five-fold validation, the selected category-level setting is $\lambda=0$, and the selected task-level setting is $(\lambda,\mu)=(0,0.03)$.
The final benchmark rubrics are refitted on all retained preferences using the setting selected in five-fold validation, with feature scaling and numerical optimization detailed in Appendix~\ref{sec:optimization-details}.

\begin{figure*}[!htb]
    \centering

    \begin{subfigure}[t]{0.46\linewidth}
        \centering
        \includegraphics[width=\linewidth]{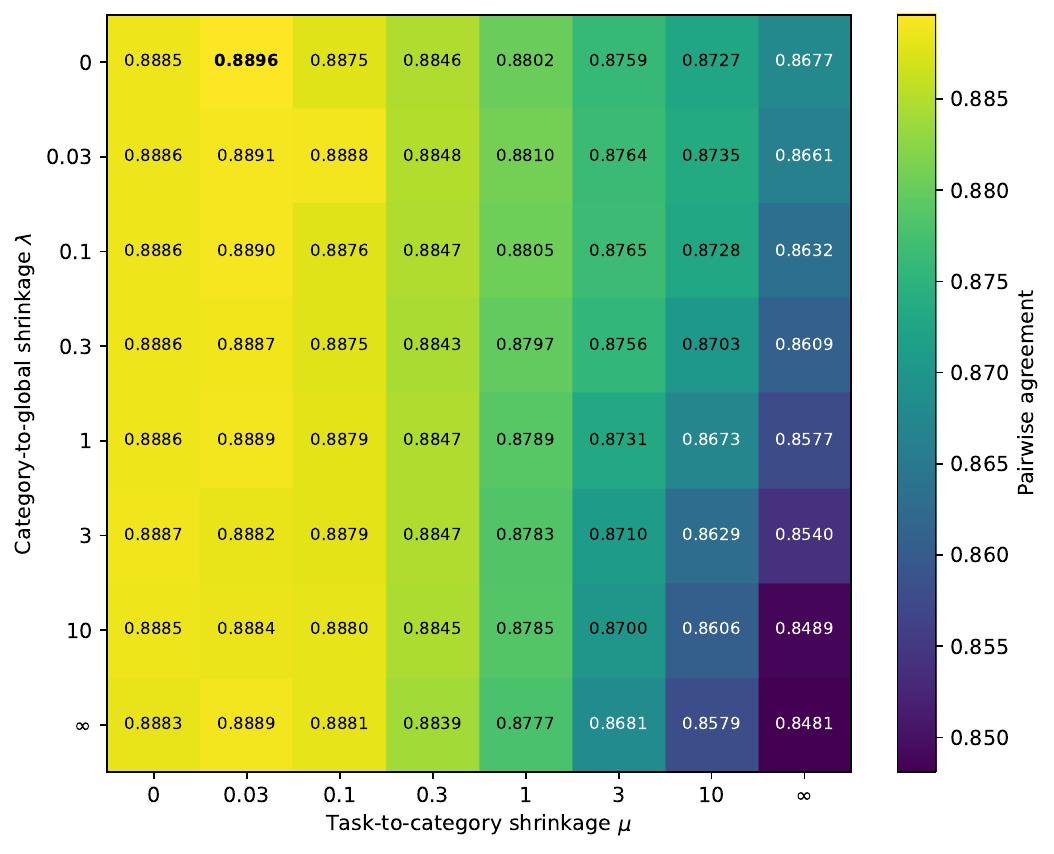}
        \caption{Joint sensitivity}
        \label{fig:weight-sensitivity-joint}
    \end{subfigure}
    \hfill
    \begin{subfigure}[t]{0.46\linewidth}
        \centering
        \includegraphics[width=\linewidth]{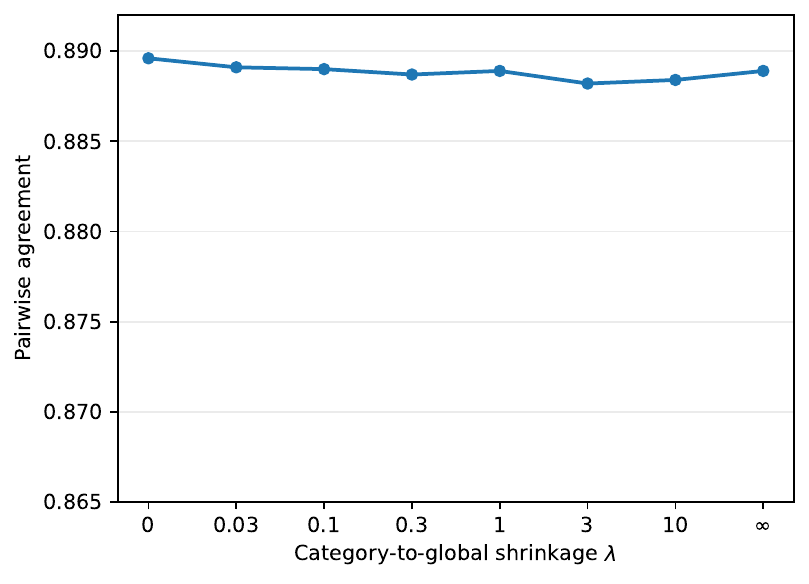}
        \caption{Fixed $\mu=0.03$}
        \label{fig:weight-sensitivity-lambda}
    \end{subfigure}
    \par\medskip

    \begin{subfigure}[t]{0.46\linewidth}
        \centering
        \includegraphics[width=\linewidth]{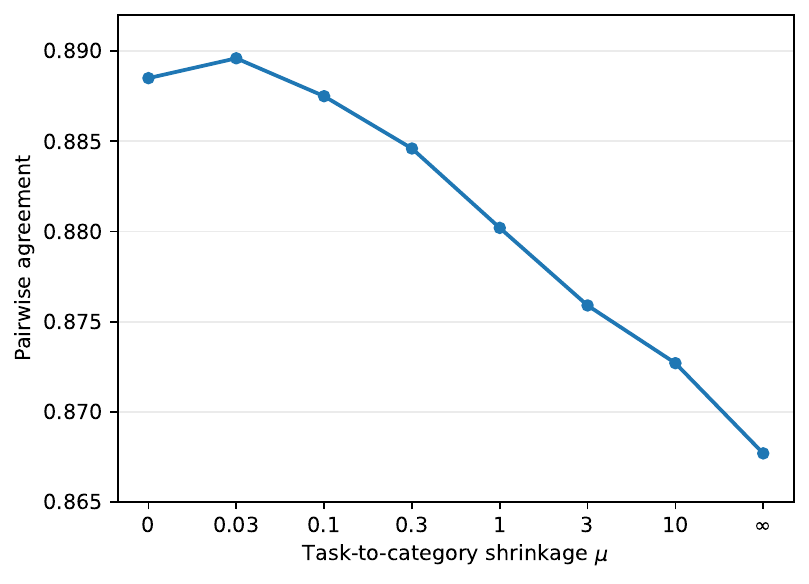}
        \caption{Fixed $\lambda=0$}
        \label{fig:weight-sensitivity-mu}
    \end{subfigure}
    \hfill
    \begin{subfigure}[t]{0.46\linewidth}
        \centering
        \includegraphics[width=\linewidth]{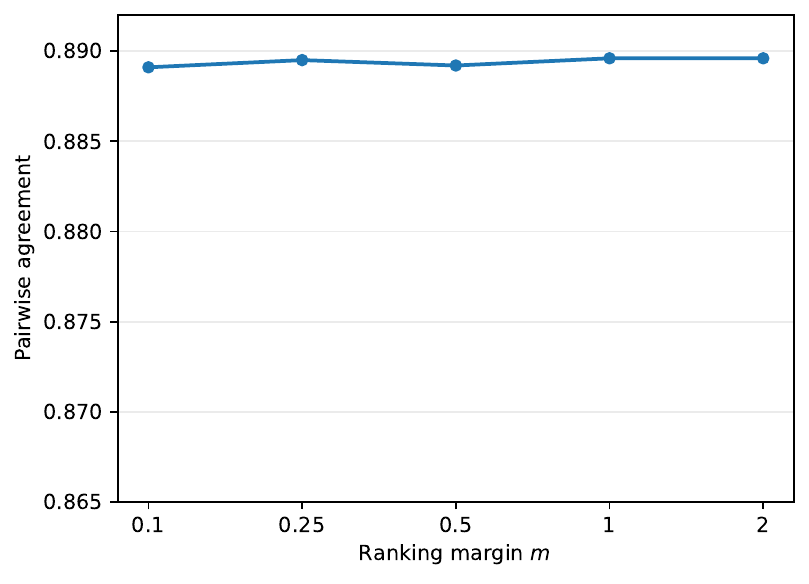}
        \caption{Ranking margin $m$}
        \label{fig:weight-sensitivity-margin}
    \end{subfigure}

    \caption{
    Sensitivity of weight learning to the hierarchical shrinkage coefficients
    $\lambda$ and $\mu$ and the ranking margin $m$.
    Panel (a) reports held-out pairwise agreement over the full
    $8\times8$ $\lambda$--$\mu$ grid.
    Panels (b) and (c) vary one shrinkage coefficient while fixing the other
    at its selected value.
    Panel (d) varies the ranking margin $m$ with an auxiliary coefficient-share cap $\rho=0.5$.
    The value $10^6$ is displayed as $\infty$.
    }
    \label{fig:weight-sensitivity}
\end{figure*}

Figure~\ref{fig:weight-sensitivity-joint} shows agreement over the full $8\times8$ search grid; Figures~\ref{fig:weight-sensitivity-lambda} and~\ref{fig:weight-sensitivity-mu} show slices through the selected setting.
The selected task-level setting achieves 0.8896 agreement.
With $\mu=0.03$ fixed, varying $\lambda$ gives agreement between 0.8882 and 0.8896, a range of 0.14 percentage points.
With $\lambda=0$ fixed, varying $\mu$ gives 0.8677--0.8896.
Agreement is more sensitive to task-level sharing ($\mu$) than to category-level sharing ($\lambda$), while remaining stable near the selected setting.

\paragraph{{Sharing under limited annotation.}}
{The selected $\lambda=0$ reflects the amount of annotation available: every category has enough preference pairs to be fitted without shrinkage toward the global weights.
To examine sharing when annotation is scarce, we retain only a fraction of the training queries of every task, refit the hierarchy, and evaluate on the held-out queries (Table~\ref{tab:low-annotation}).
With fewer pairs, larger sharing coefficients are selected, and the gain of the hierarchy over task weights fitted independently grows from 0.4 percentage points with all annotations to 5.6 points with 5\%.
At every level of annotation, the hierarchical task weights are more accurate than the global weights.}

\begin{table}[htb]
\centering
\small

\caption{{
Held-out agreement when only a fraction of the training queries of each task is retained, averaged over three random subsamples.
\emph{Independent} fits each task without a parent; \emph{Hierarchical} uses the selected sharing coefficients.
}}
\label{tab:low-annotation}
\begin{tabular}{cccccc}
\toprule
Retained queries & Pairs per task & Global & Independent & Hierarchical & Selected $(\lambda,\mu)$ \\
\midrule
5\% & 11 & 0.8345 & 0.7845 & \textbf{0.8404} & $(3, 3)$ \\
10\% & 21 & 0.8450 & 0.8041 & \textbf{0.8536} & $(1, 1)$ \\
25\% & 52 & 0.8511 & 0.8400 & \textbf{0.8662} & $(3, 1)$ \\
50\% & 105 & 0.8506 & 0.8599 & \textbf{0.8726} & $(1, 0.3)$ \\
\bottomrule
\end{tabular}
\end{table}

\subsubsection{Sensitivity to the Ranking Margin}
\label{sec:margin-ablation}

The margin diagnostic evaluates $m\in\{0.1,0.25,0.5,1,2\}$ at $(\lambda,\mu)=(0,0.03)$, with an auxiliary coefficient-share cap $\rho=0.5$.
Figure~\ref{fig:weight-sensitivity-margin} shows this fixed-$\rho$ slice of the $5\times5$ sweep over those margins and $\rho\in\{0.2,0.3,0.4,0.5,1\}$.
All cells reuse the same folds and the feature scales estimated from each fold's training queries.
Agreement ranges from 0.8891 to 0.8896, a range of 0.05 percentage points.
The auxiliary capped diagnostic therefore shows little variation across the tested margins.

\FloatBarrier

\subsubsection{Ability-Scoring Judge Reliability}
\label{sec:ability-judge-reliability}

Using responses from seven model configurations, we evaluate consistency among the three judges at both the response--ability level and the model-ranking level.
This analysis uses the single-ability scoring protocol described in Appendix~\ref{sec:judge-details}.

Table~\ref{tab:judge-instance-agreement} reports pairwise agreement between
judge scores.

\begin{table}[!h]
\centering
\caption{
Pairwise agreement between ability-scoring judges.
MAD denotes mean absolute difference, and Within-1 denotes the fraction of
scores differing by at most one point.
}
\label{tab:judge-instance-agreement}
\begin{tabular}{lcccc}
\hline
Judge pair & MAD & Within-1 & Spearman & Weighted $\kappa$ \\
\hline
Gemini--Gemma
& 0.75 & 86.9\% & 0.65 & 0.63 \\

Qwen--Gemma
& 0.80 & 85.9\% & 0.53 & 0.56 \\

Qwen--Gemini
& 1.08 & 82.1\% & 0.49 & 0.51 \\
\hline
\end{tabular}
\end{table}

Ranking these configurations by each judge's mean response--ability score yields higher agreement.
The pairwise Spearman correlations are 0.93 for Gemini--Gemma,
0.89 for Qwen--Gemma, and 0.82 for Qwen--Gemini.
The corresponding Kendall correlations are 0.81, 0.81, and 0.62.

Judge disagreement is greater for subjective criteria (mean MAD 1.777) than for objective LLM-judged criteria (0.959), indicating that interpretive judgments account for more of the variation among evaluators.

\paragraph{{Results with a single judge.}}
{Table~\ref{tab:per-judge-agreement} repeats the preference-agreement analysis with ability scores from one judge at a time, under the protocol of Table~\ref{tab:weight_granularity2}.
For every judge, agreement increases from conventional LLM-as-a-judge scoring to uniform ability aggregation, and then from global to category-level to task-specific weights.
The model ranking is also stable across judges: the overall scores in Table~\ref{tab:model-family-main} recomputed with a single judge have Spearman correlations of 0.94 to 0.99 with the three-judge ranking, and the same four configurations lead under every judge.}

\begin{table}[htb]
\centering
\small

\caption{{
Agreement with human preferences when the ability scores come from a single judge.
}}
\label{tab:per-judge-agreement}
\begin{tabular}{lccccc}
\toprule
Judge & Conventional & Uniform & Global & Category & Task \\
\midrule
Gemini-3.1-Pro & 0.7168 & 0.7546 & 0.7949 & 0.8134 & 0.8306 \\
Qwen3.6-35B-A3B & 0.6750 & 0.7329 & 0.7784 & 0.8010 & 0.8196 \\
Gemma-4-31B & 0.6523 & 0.7435 & 0.7789 & 0.7924 & 0.8058 \\
\bottomrule
\end{tabular}
\end{table}

\end{document}

%% file: math_commands.tex
\usepackage{amsmath,amsfonts,bm}

\def\eqref#1{equation~\ref{#1}}

\def\1{\bm{1}}

\DeclareMathAlphabet{\mathsfit}{\encodingdefault}{\sfdefault}{m}{sl}
\SetMathAlphabet{\mathsfit}{bold}{\encodingdefault}{\sfdefault}{bx}{n}

